\documentclass[a4paper,fleqn]{cas-sc}

\usepackage[numbers]{natbib}
\usepackage{float}
\usepackage{tabularx}
\def\tsc#1{\csdef{#1}{\textsc{\lowercase{#1}}\xspace}}
\tsc{WGM}
\tsc{QE}
\begin{document}
\let\WriteBookmarks\relax
\def\floatpagepagefraction{1}
\def\textpagefraction{.001}
%\let\printorcid\relax 

% \shorttitle{<short title of the paper for running head>}
%\shorttitle{Leveraging social media news to predict stock index movement using RNN-boost}

% \shortauthors{<short author list for running head>}
%\shortauthors{V. {{\=A}}nand Rawat et al.}

\title[mode = title]{Industrial brain: a human-like autonomous cognitive decision-making and planning system}

% \tnotemark[<tnote number>]
%\tnotemark[1,2]

%\tnotetext[1]{This document is the results of the research project funded by the National Science Foundation.}
%\tnotetext[2]{The second title footnote which is a longer text matter to fill through the whole text width and overflow into another line in the footnotes area of the first page.}


% \author[<aff no>]{<author name>}[<options>]
\author[1,2]{Junping Wang}
\cormark[1]
\cortext[1]{Junping Wang}
\ead{junping.wang@ia.ac.cn}

\author[1,2]{Bicheng Wang}
\ead{wangbicheng2022@ia.ac.cn}

\author[1,2]{Yibo Xue}
\ead{xueyibo2023@ia.ac.cn} 

\author[3]{Yuan Xie}
\ead{yxie@cs.ecnu.edu.cn} 

\address[1]{State Key Laboratory of Multimodal Artificial Intelligence Systems, Institute of Automation, Chinese Academy of Sciences, Beijing, China}
\address[2]{School of Artificial Intelligence, University of Chinese Academy of Sciences, Beijing, China}
\address[3]{School of Computer Science and Technology, East China Normal University, Shanghai, China}
% \cormark[1]

% \fnmark[<footnote mark no>]

% \ead{<email address>}

% \ead[url]{<URL>}

% \credit{<Credit authorship details>}

% \affiliation[<aff no>]{organization={},
% %          citysep={}, % Uncomment if no comma needed between city and postcode

% \author[<aff no>]{<author name>}[<options>]

% Footnote of the second author

% Email id of the second author
% \ead{}



% \affiliation[<aff no>]{organization={},
% %          citysep={}, % Uncomment if no comma needed between city and postcode

\cortext[1]{Corresponding author}



%\author[1,3]{V. {{\=A}}nand Rawat}[type=editor,auid=000,bioid=1,prefix=Sir, role=Researcher, orcid=0000-0001-7511-2910]
%\cormark[1]
%\ead{cvr_1@tug.org.in}
%\ead[url]{www.cvr.cc,www.tug.org.in}
%\credit{Conceptualization of this study, Methodology, Software}

%\author[2,4]{Han Theh Thanh}[style=chinese]

%\author[2,3]{T. Rishi Nair}[role=Co-ordinator, suffix=Jr]
%\ead{rishi@sayahna.org}
%\ead[URL]{www.sayahna.org}
%\credit{Data curation, Writing - Original draft preparation}

%\author[1,3]{Karl Berry}
%\cormark[2]
%\fnmark[1,3]
%\ead{karl@freefriends.org}
%\ead[URL]{www.tug.org}

%\address[1]{Indian \TeX{} Users Group, Trivandrum 695014, India}
%\address[2]{Sayahna Foundation, Jagathy, Trivandrum 695014, India}
%\address[3]{\TeX{} Users Group, Providence, MA, USA}

% \cortext[2]{Principal corresponding author}

\begin{abstract}
Resilience non-equilibrium measurement, the ability to maintain fundamental functionality amidst failures and errors, is crucial for scientific management and engineering applications of industrial chain. The problem is particularly challenging when the number or types of multiple co-evolution of resilience (for example, randomly placed) are extremely chaos. Existing end-to-end deep learning ordinarily do not generalize well to unseen full-feld reconstruction of spatiotemporal co-evolution structure, and predict resilience of network topology, especially in multiple chaos data regimes typically seen in real-world applications. To address this challenge, here we propose industrial brain, a human-like autonomous cognitive decision-making and planning framework integrating higher-order activity-driven neuro network and CT-OODA symbolic reasoning to autonomous plan resilience directly from observational data of global variable. The industrial brain not only understands and model structure of node activity dynamics and network co-evolution topology without simplifying assumptions, and reveal the underlying laws hidden behind complex networks, but also enabling accurate resilience prediction, inference, and planning. Experimental results show that industrial brain significantly outperforms resilience prediction and planning methods, with an accurate improvement of up to 10.8\% over GoT and OlaGPT framework and 11.03\% over spectral dimension reduction. It also generalizes to unseen topologies and dynamics and maintains robust performance despite observational disturbances. Our findings suggest that industrial brain addresses an important gap in resilience prediction and planning for industrial chain.
\end{abstract}



% Keywords

\maketitle

% Main text
\section{Introduction}\label{1}
The industrial chain\cite{Teng2017analysis} is as giant high-order interactions network with large-scale interconnected nodes and weighted links, which is crucial for the healthy and orderly development of the national economy. The co-evolution chaotic characteristics lead to the emergence of complex multiple percolation phenomena \cite{Sun2021higher}, articulating led to non-equilibrium percolation dynamical disasters \cite{Bianconi2024theory} in industrial chain. 
To address the complex multiple percolation issues, the concept of industrial chain 
resilience\cite {Gao2016Universal} was formally defined recently, articulating that a resilient system should invariably
converge to a desired stable equilibrium after perturbation. Prior research indicates that the loss of resilience often has extensive implications. Typically, the resilience measurements follows the principle of multiple equilibria , which is manifested in two specific scenarios: (1) non-trivial stable equilibrium of node activities, indicating that node status equilibria are not measured after being disturbed or disrupted, leading to challenges for node dynamic reconstruction with unstructured data, (2) non-trivial stable equilibrium of network co-evolution topology, where spatiotemporal dynamics is generalized in a dynamical system with multiple percolation. For example, the co-evolution chaotic characteristics of industrial chain, lead to the emergence of multiple percolation phenomena, which will trigger catastrophic resilience. This underscores the profound consequences of resilience loss and highlights the urgent need for approaches to predict and mitigate such outcomes.

Existing network resilience methods provides analytical estimates for resilience of \(N\)-dimensional systems with complex interactions between components by reducing them to tractable one dimensional systems based on mean-field theory, spectral graph theory and ResInf\cite {2024Deep}, etc. 
Despite their groundbreaking successes, these approaches often make strong assumptions equilibria about network topology and dynamics for analytical feasibility. One success story is that of bayesian cognitive models \cite{griffiths2010probabilistic}, which optimally combine domain prior knowledge with bayesian network. These developments occurred in interaction with statistics and machine learning, where a unified perspective on probabilistic empirical inference has emerged. This model provides representation and reasoning theory 
for network resilience reconstruction. Another importance advance is that of human-level cognitive control model with large state spaces \cite{Wang2018multitask}, where all physical object operating as part of the distributed 
coordination cognitive computing network will run in an iterative three-step automate optimizing closed loop : learning control policies from the environment, making decisions based on present state, and responding by performing actions that affect the environment \(-\) either by altering operational status or via communication with other objects. The method provides a significant theory for planning resilience multiple equilibria. 

\begin{figure}[!htbp]
	\centering
	\includegraphics[width=16cm]{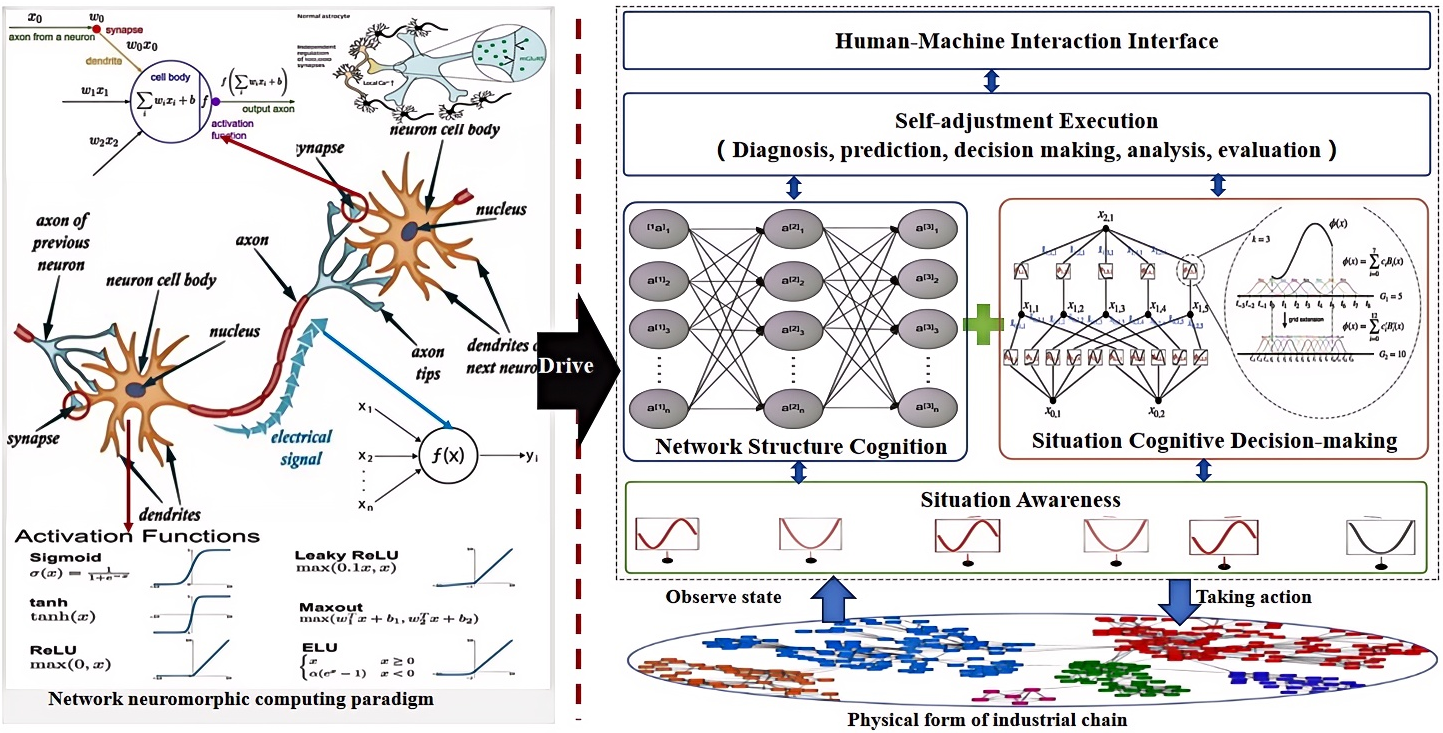}
	\caption{The overall framework of neuro-symbolic autonomous measurement. Simulating intuitive nervous system and logical reasoning system of human brain, the framework defines human-level autonomous measurement close-loop for awareness, cognition, decision-making, and self-adjustment of resilience non-equilibrium. Thereby effectively maintaining the resilience and robustness of the entire industrial chain network.}\label{fig7}
\end{figure}

Unfortunately, the resilience of industrial chain is a complex and chaotic percolation transition system, frequently in a state of non-equilibrium, exhibiting complex dynamic characteristics, limiting their applicability to real-world systems. The complex dynamics system of the industrial chain is like to the human brain’s nervous system. Network neuroscience\cite{Bassett2017network}, which is used to characterize the structure, function, and predictive dynamics of the human brain, offers a novel research approach to address the resilience of non-equilibrium dynamics in industrial chains. In other words, how to construct resilience non-equilibrium measurements on industrial chains using network neuroscience methods has become a current research focus in the field of artificial intelligence\cite{Beaty2019network}, as illustrated in Figure \ref{fig7}. The specific new task requirements without the mentioned capabilities are as follows:
\begin{itemize}
    \item[1)] \textbf{Situation awareness:} The industrial chain is a complex giant system composed of large-scale personnel, systems, and environmental heterogeneous element nodes, which high-order interact dynamically in multiple domains. This makes it difficult for existing deep learning methods to effectively perceive the evolution of large-scale dynamic situations. The section is an urgent need for a situational awareness system, providing a multi-scale intuitive perception capability like to human neuron, obtain the evolution information of large-scale personnel, systems, and environmental heterogeneous nodes, identify potential risks, and predict the activity situation of risk factors affecting the resilience of the industrial chain operation. 
    \item[2)] \textbf{Cognitive thinking system:} Considering that the industrial chain is a massive multi-domain spatio-temporal multi-dimensional interactive chaotic network system, modeling the organizational structure of such a high-order complex network system and achieving virtual-intelligent interaction becomes a research challenge in the field of artificial intelligence. This requires the industrial brain to provide a logical thinking system similar to the central nervous system of the human brain. Referring to human brain histology, the goal is to achieve hierarchical and domain-based network modeling of large-scale element nodes and network topology structures within the industrial chain, and map them into the cognitive thinking system of the industrial brain. This facilitates deep understanding of the behaviors of large-scale elements in the industrial chain, and meets the needs for comprehensive cognitive reasoning analysis of the resilience of the industrial chain’s operation from macro, meso, and micro perspectives.
    \item[3)] \textbf{Cognitive decision-making}: The operation of the industrial chain is subject to interference from internal and external uncertainties, leading to rapid evolution or even catastrophic changes in resilience. How to use artificial intelligence technology to comprehensively assess the catastrophic changes in the situation has become a current research challenge \cite{Lynn2019physics}. This requires the industrial brain to provide an autonomous cognitive decision-making system similar to the human brain. Under human supervision, it can accurately detect such operational catastrophic changes. Through a combination of virtual and real deduction and evaluation methods, it achieves a comprehensive assessment of the trend of situation evolution, predicts potential risks, generates response plans, and improves the accuracy and effectiveness of future resilience situation evolution decisions.
    \item[4)] \textbf{Resilience self-adjustment}: Considering that the resilience of the industrial chain is akin to a complex network percolation and transformation phenomenon, it necessitates that the industrial brain offers a regulation execution system similar to the human brain. This system is tasked with detecting potential risks and vulnerabilities within the industrial chain, issuing early warnings proactively, and strategically reallocating resources within the chain to enhance its self-adaptation to changing environmental conditions. Additionally, it must self-repair its resilience to accommodate the growth and evolution of the industrial chain.
\end{itemize}

In order to meet the above requirements, this paper proposes a human-like autonomous cognitive decision-making and planning framework to handle the resilience catastrophe, which we call industrial brain. The proposed industrial brain can combine autonomous topological deep Learning and automated symbolic reasoning, and is easy to construct new "human-in-the-loop" neuro-symbolic hybrid augmented intelligent system in artificial intelligence field. 
First, we employ neuro-symbolic cognitive thought large model in industrial brain to better capture the underlying high-order correlations among data, the neuro-symbolic cognitive thought large model is used for giant chaotic network modeling, which encodes node activity dynamics and co-evolutionary topologies with cellular automatas. Each cognitive cellular automatas in a graph can only connect several isomorphic nodes, while in a cognitive functional network, each functional group can connect more than two cognitive cellular automatas and thus is more flexible. When handling chaotic network data, the cognitive thought large model can generate different types of functional group using the multi-type topology data and then directly concatenates these functional group into one cognitive functional network.
Second, we develop "human-in-the-loop" autonomous interactive decision-making model, where 
autonomously perceive the evolutionary trends of various nodes in giant chaotic networks,
understand the physical mechanisms of the continuous evolution of the current situation within a specific space-time context, automatically predict the potential causal factors for catastrophic events at specific times, and conduct iterative simulations and evaluation calculations that integrate virtual and real scenarios of the prediction results. This aims to resolve conflicts in decision-making strategies, generate the optimal resilience structure planning scheme, and autonomously adjust the resilience of the industrial chain structure. Consequently, this constructs an autonomous cognitive decision-making closed-loop control computational mechanism to automatically mitigate major events that disrupt the resilience of industrial chain. 

The main contributions of industrial brain can be summarized as the follows:
\begin{itemize}
	\item[1)] Drawing on the cognitive computational mechanisms of the human brain, we have defined the concept and technical framework of a human-like industrial brain. Under human supervising, it not only provides various cognitive measurement functions for events affecting resilience mutations, such as perception, reasoning, and decision-making, but also can accurately detect, understand, analysis and control multi-scale nonlinear dynamical catastrophic in industrial chain, preventing resilience catastrophe in industrial chain. 
	
	\item[2)] We construct neuro-symbolic cognitive thought large model of industrial brain via combining dynamical neural networks and knowledge-guided symbolic reasoning\cite{Sarker2021neuro} from high-order interaction data of industrial chain. This diagram of thought large model provides a high-order interaction modeling tool, efficiently maps multi-scale spatiotemporal co-evolution organization into hierarchical dynamic neural network.  It also enable the industrial brain to possess human-like cognitive diagram of thought abilities for understand and performing various cognitive tasks.
	
	\item[3)] We develop human-level autonomous CT-OODA running engine for industrial brain. The engine employs awareness, cognition, decision-making, and planning operator into autonomous CT-OODA closed-loop, established the human-like autonomous decision-making in industrial brain for handling various resilience evolution event. It provides many capabilities of autonomous cognition, decision, planning for resilience disaster of industrial chain, facilitating accurate prevention of resilience disasters with linear classifiers. 
	
	\item[4)] Extensive experiments on specific network data scenario of car auto parts industry chain, are conducted to demonstrate the effectiveness, prediction accuracy, and robustness of the proposed method. Besides, detailed mathematical discussions are provided to offer a deeper understanding of the proposed industrial brain, the setting of performance comparison criteria and the discussion and analysis of experimental results.
	
\end{itemize}

The remainder of the paper is organized as follows: Section \ref{2} introduces related works of industrial brain. Section \ref{3} describes the proposed autonomous cognitive and decision-making framework in detail, combines the fundamental theories of autonomous cognitive neural networks, neuro-symbolic learning and reasoning, and adaptive intelligent control to construct the industrial brain’s active thought, autonomous cognitive decision-making capabilities. Experimental analysis and completion results are shown in Section \ref{4} to verify our method. Finally, we conclude the proposed method in Section \ref{5}.

\section{Related Work}\label{2}

The industrial brain is driving by cognitive computational neuroscience, has become the main research direction in the field of artificial intelligence. Significant progress has been made, primarily focusing on diagram of thought modeling, cognitive neural computing and neuro dynamic regulation.

\subsection{Cognitive diagram of thought of the brain}

Neural thought language modeling\cite{Besta2024graph} is an important branch of network neuroscience, which aims to simulate human thought process and language understanding ability. Through the combination of neural network models and cognitive architecture, neural thought language modeling can achieve more intelligent and humanized language processing capabilities, including cognitive tasks such as understanding language context, reasoning and decision-making. As a landmark method in the field of neural thought language modeling, Chain-of-Thought (CoT) \cite{Wei2022chain} is able to model the information generated by large language models (LLMs) as arbitrary graphs, in which the information units are the nodes of the large language model, and the edges correspond to the dependencies between these nodes. This approach makes it possible to combine arbitrary LLMs thoughts into synergistic outcomes, extract the essence of an entire network of thoughts, or use feedback loops to augment thoughts. The Uncertainty-based Active Learning method\cite{Wang2018certainty} is used to make the most uncertain prediction that CoT has the ability to actively recognize the external environment, and then manually annotate these specific thought language types of questions to improve the performance of neural thought language models.

Friston. et al\cite{Friston2006free} proposed the principle of free energy model and constructed a hierarchical neural thought language construction method. Aguilera, M.et al combined the advantages of the principle of free energy and predictive coding to propose a dynamic cognitive thought model (Auto-CoT)\cite{Aguilera2022particular} , which can express the basic functions of neural thought language and automatically suggest enhancement and selection, thereby reducing the need for manual annotation. The Voting Strategy method\cite{Myatt2007theory}improves the linguistic reasoning ability of neural thought, generating multiple answers using different reasoning paths, and then selecting the most consistent answer as the final output. The Self-Taught Reasoner method STaR (Self-taught Reasoner)\cite{Zelikman2022star} helps neural mental language models fine-tune the reasoning that ultimately produces the correct answer. In-context Learning\cite{Dong2022survey}, which provides an approach to problem solving in the context of a new task through individual prompts, allows the model to effectively learn how to solve each part before combining them to solve the main task. These methods represent a shift towards more efficient and effective use of LLMs, enabling them to perform a wider range of tasks without extensive retraining or fine-tuning. They also demonstrate the potential of combining human reasoning and the powerful pattern recognition capabilities of LLMs, a key area of AI research and development.

\subsection{Cognitive decision-making computing of the brain}
Cognitive decision-making computing\cite{Kriege2018cognitive} is an important research field of artificial intelligence. It aims to simulate the cognitive computing mechanism of human brain, develop cognitive computing models and technologies, simulate the cognitive thought of human brain, and make the intelligence level of machine closer to human intelligence level. Dr. H. Andrew Schwartz et al. developed the SOAR cognitive neural computing framework\cite{Butt2013soar}, which integrates problem solving, learning and knowledge representation into the cognitive computing framework. By creating detailed mathematical models and simulations, it simulates the process of human cognition, so that intelligent systems can deal with complex tasks and environments like humans. Although SOAR has unique contributions and applications in the field of cognitive neural computing, it also has limitations such as dependence on human prior rule support, difficulty in handling large amounts of data, lack of natural language processing ability, lack of adaptability and generalization ability, and lack of multi-modal perceptual integration. It provides valuable insights and foundation for subsequent research on cognitive neural computing.

Another influential cognitive neural computing architecture, the Adaptive Control of Thought-Reasoner (ACT-R)\cite{Ritter2019act}, was developed by researchers such as John Anderson. It is based on the principles of cognitive psychology and cognitive science, simulates human behavior in solving cognitive tasks, and establishes cognitive neural computing architecture with perception, memory, reasoning, language and planning. These modules interact through a brain computer interface of working memory. ACT-R has been applied in many fields, including education, artificial intelligence, and human-computer interaction. In education, ACT-R's model can help in the design of intelligent tutoring systems that are able to provide personalized instructional support based on the learning habits and abilities of students. In artificial intelligence research, ACT-R provides a computational framework to simulate human cognition, which helps to develop more intelligent and adaptive software systems. With the advancement of neuroscience and computing technology, cognitive neural computing architectures such as ACT-R will continue to be improved to better simulate human cognitive processes and play a greater role in fields such as artificial intelligence and cognitive assistive technologies. Although ACT-R has opened up the research direction of autonomous reasoning cognition in the field of cognitive neural computing, it also has some limitations, such as dependence on a large number of rule support, weak domain generalization ability, lack of natural language representation and processing, lack of interpretability and transparency. It provides valuable insights and foundation for subsequent research on cognitive neural computing.

DUAL (Distributed Unit for Assembling Learning)\cite{Simonivc2019autonomous} is a hybrid cognitive neural computing architecture developed by the Cognitive Neuroscience Laboratory of the Institute of Psychology, Chinese Academy of Sciences. It consists of two main components: DEDUCTER (deductive reasoning component) and ACTER (rule-based action planning and decision component). DEDUCTER uses a method called "logic programming" to simulate the human reasoning process by writing and executing logical rules. ACTER uses a method called "decision table" to simulate the human decision process by writing and executing decision rules. The strength of the DUAL framework lies in its ability to model human reasoning and decision-making processes, which can be used to model human cognitive behavior and help develop intelligent systems. In practice, the DUAL architecture can be used to study the cognitive mechanisms of people when performing various tasks, such as decision-making and behavior in complex environments such as driving, aviation operations, and educational learning. 

A novel intelligent framework, referred to as OlaGPT\cite{Xie2023olagpt}, which carefully studied a cognitive architecture framework, and propose to simulate certain aspects of human cognition. The framework involves approximating different cognitive modules, including attention, memory, reasoning, learning, and corresponding scheduling and decision-making mechanisms. Inspired by the active learning mechanism of human beings, it proposes a learning unit to record previous mistakes and expert opinions, and dynamically refer to them to strengthen their ability to solve similar problems. The paper also outlines common effective reasoning frameworks for human problem-solving and designs Chain-of-Thought (COT) templates accordingly. A comprehensive decision-making mechanism is also proposed to maximize model accuracy. The efficacy of OlaGPT has been stringently evaluated on multiple reasoning datasets, and the experimental outcomes reveal that OlaGPT surpasses state-of-the-art benchmarks, demonstrating its superior performance.

Goel. et al\cite{goel2024neurosymbolic} proposed A neurosymbolic cognitive architecture framework for handling novelties in open worlds.This framework combines symbolic planning, counterfactual reasoning, reinforcement learning, and deep computer vision to detect and accommodate novelties. The ability to detect and accommodate novelties allows agents built on this framework to successfully complete tasks despite a variety of novel changes to the world. Extensive multi-stage evaluations of an architecture instance of the framework performing a crafting task in a Minecraft-inspired simulation environment demonstrated that the proposed methods can usually detect, to some extent characterize, and often accommodate several types of novelties, such as novel objects or novel agents.  These encouraging results point the way for future developments. Possible improvements could include the addition of other subsymbolic input modalities like sound or tactile sensors, extended symbolic inference and novelty exploration mechanisms that use planners to plan experiments for discovery instead of fixed strategies, and yet better ways for quickly experimenting with objects and actions in the environment to discover solutions to unforeseen and unexpected problems. 

These systems can flexibly adjust the proportion of automation and control processing according to user's needs and environmental changes, so as to provide more humane and efficient services, so as to better simulate and understand human cognitive processes, and provide new theoretical basis and technical support for the development of artificial intelligence. However, the four framework has limitations of cognitive ability in dealing with large-scale and complex decision rule learning, reasoning, and memorization. 

\subsection{Neural network dynamics regulation and control of the brain}

Neural network dynamics regulation and control \cite{Luan2023neurodynamic} is an important branch of network neuroscience. Its core idea is to use the dynamic characteristics of the nervous system to regulate and control the behavior and physiological process of the organism. The nervous system\cite{Plata1991immunoregulators} is a highly complex dynamic system in organisms. It regulates various behaviors of organisms through the interaction and information transmission between neurons. The early Isching-like model pioneered the regulation of neural dynamics. The data-driven neural dynamics model\cite{Freestone2011data} constructed by Maximum Entropy Techniques can make long-range correlation predictions in neurons and brain regions, and is used to explain critical or avalanche behavior in neuronal systems. As a landmark technology of neurodynamic regulation, non-invasive stimulation technology\cite{Guo2023novel} is introduced into the network model of brain dynamic process, which realizes the stimulation for specific diseases, and adjusts the strategy according to the neural activity after stimulation, and realizes the feedback closed-loop regulation of neuronal excitability, neural network connectivity, neural signal processing, learning and memory, and brain function regulation.

Researchers began to develop Neural Mass Models based on Wilson-Cowan's population dynamics model\cite{Verma2024dynamics}, which accurately described the mean field of large-scale neural activity in the brain. To facilitate the process of simulation, the Kuramoto model of the oscillatory dynamic process is used as a simplified neural mass model to provide insight into the stochastic synchronization of neural oscillations. The application of neurodynamic regulation is very wide. It is not only used in diseases and abnormalities of the human nervous system, such as epilepsy, Parkinson's disease, etc., but also can be used to develop advanced scientific and technological products such as intelligent transportation, industrial brain and urban brain. This paper provides technical ideas for the research of industrial brain timely intervention in the percolation phase change dynamics of industrial chain. Researchers have proposed the uplink loop system as a neuroregulatory control method for complex adaptive dynamics in the brain, giving the industrial brain flexibility, robustness and efficiency, and preventing major disasters such as blocking points, breaking points, leakage points and blind spots of the industrial chain.

\begin{figure}[!htbp]
	\centering
	\includegraphics[width=16cm]{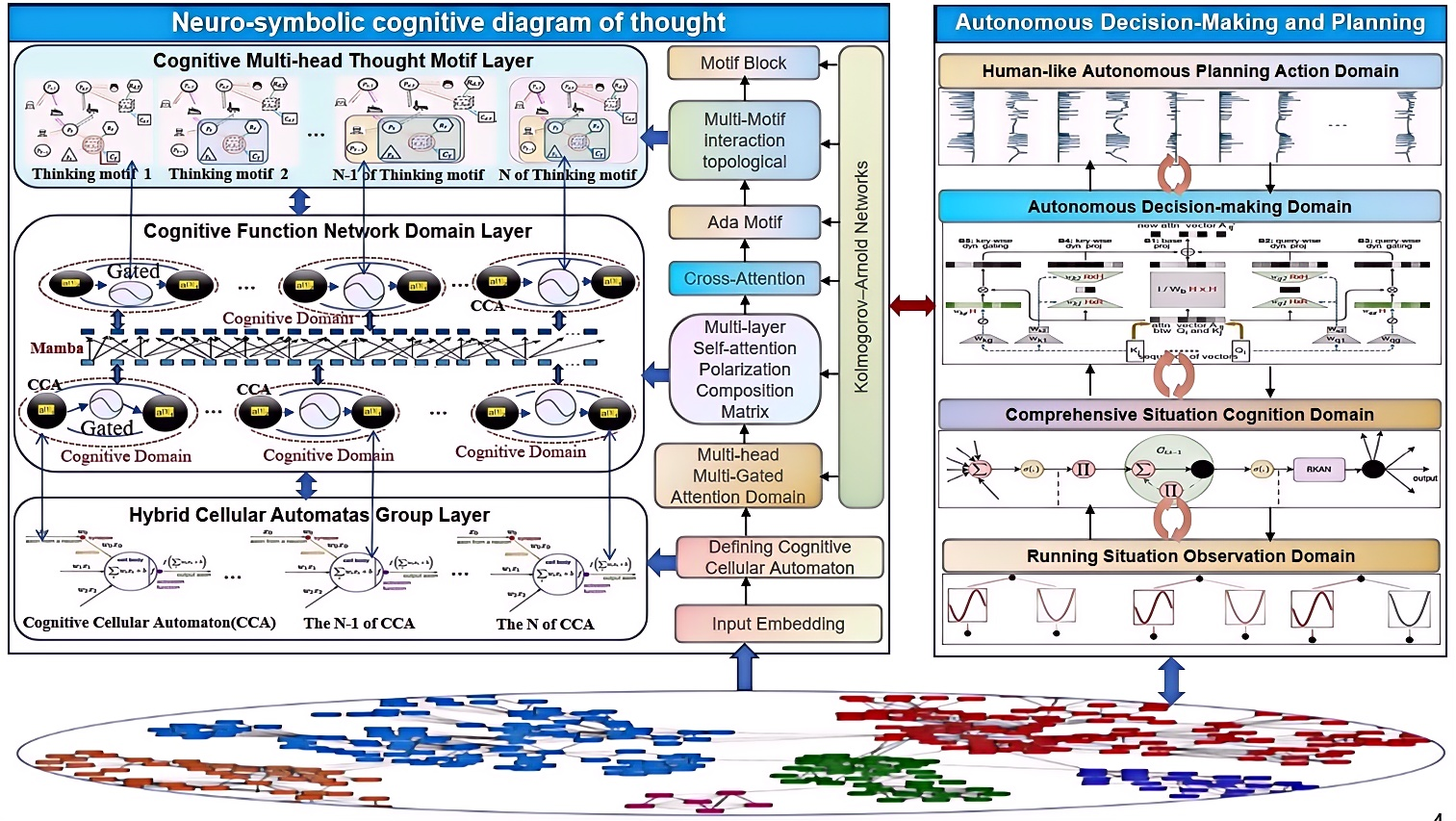}
	\caption{The proposed industrial brain framework. industrial brain leverages simulated the human brain`s intuitive cognitive system and logical analysis systems to prevent the resilience disasters events of industrial chain without any prior assumption. It contains two modules: neuro-symbolic cognitive diagram of thought module generates neuro-symbolic cognitive thought large model from large-scale innovation elements in the industrial chain. We map industrial chain nodes activity status to dense representations with cognitive thought function group layers, which models the complex correlations between node activities and produces representations for node activity dynamics. Multiple nodes co-evolution topology are aggregated to cognitive thought function domain layers, to generate discriminating topological representations for each node’s multi-hop neighborhoods,k-space projector aggregates the representations for node activity dynamics and topologies via a virtual global node. Cognitive thought interaction motif layers provide autonomous messaging interaction channel among multiple virtual global node, employs a multi-head self-attention network domain to fuse the representations learned from various trajectories dynamically. Another autonomous decision-making module is integrating situation autonomous awareness, comprehensive situation cognition, co-evolution planning decision-making, and co-evolution topology planning action into the autonomous computing CT-OODA loop, providing capabilities of autonomous cognition, learning, and reasoning for industrial brain in a multi-object dynamic network environment, facilitating accurate catastrophic resilience identification and control with industrial brain.}\label{fig8}
\end{figure}

\section{Methodology}\label{3}

In this section, we briefly introduce the industrial brain, which aims to provide a human-like autonomous cognitive decision-making and planning framework for preventing resilience failures in the industrial chain. It is composed of three key components, as shown in Figure \ref{fig8}, neuro-symbolic cognitive diagram of thought, human-like autonomous decision-making, and human-like autonomous planning action.

\subsection{Neuro-symbolic cognitive diagram of thought}
In this section, we introduce how to construct a cognitive diagram of thought from giant chaotic network by neuro-symbolic joint learning framework. This model is divided into three layers: hybrid cellular automatas group layer, cognitive function network domain layer, and cognitive multi-head thought motif layer. 
The neuro-symbolic cognitive diagram of thought is used to effectively model easily interpretable industrial chain organization. Knowing which variables are important in a model’s prediction and how they are combined can be very powerful in helping people understand and trust automatic decision-making systems. To better exploit the high-order correlation among giant chaotic nodes, hybrid cellular automatas group layer aggregates the representations for node activity dynamics and co-evolution topology via a the cellular automaton\cite{Sante2010cellular}. 
The cognitive function network domain layer employs a multi-head self-attention motif to fuse the representations learned from various trajectories dynamically. Subsequently, cognitive multi-head thought motif layer uses a dimension reduction interactive network to project the aggregated representation to a 1-dimensional k-space, facilitating accurate to represent and understand resilience of industrial chain with linear classifiers. 

\subsubsection{Hybrid cellular automatas group layer}
In hybrid cellular automatas group layer, each cognitive cellular automatas unit is used to abstractly represent the behavior state of key element nodes in the industrial chain. Those key element nodes is including raw materials, markets, technology, policies, innovation, enterprises, products, logistics, information, capital, talent, and services. The specific meanings of their conceptual models are as follows:

\begin{itemize}
    \item[1)] \textbf{Raw materials:} It is the starting link of the industrial chain, with foundation, diversity and sustainability. The quality and supply stability of raw materials are crucial to the operation of the entire industrial chain.
    \item[2)] \textbf{Market:} It is the fundamental driving the operation of the industrial chain, and changes in the market directly affect the design, production, sales and price of products.
    \item[3)] \textbf{Technology:} It is an important driving force in the industrial chain, including production technology, management technology, information technology, etc. The continuous innovation of technology is an important way to improve the efficiency and added value of the industrial chain.
    \item[4)] \textbf{Policy:} It is the direction that the government has an important impact on the layout, development direction and industry standards of the industrial chain, and the guidance and support of policies can promote the healthy development of the industrial chain.
    \item[5)] \textbf{Innovation:} It is the main place for industrial innovation and transformation, with multiple functions such as research and development, achievement transformation, and talent training. Innovation can bring new market opportunities and competitive advantages.
    \item[6)] \textbf{Enterprise:} It is the core body in the industrial chain, responsible for product research and development, production, sales and service. The innovation ability of enterprises, market response speed and economies of scale are the key to the development of industrial chain.
    \item[7)] \textbf{Product:} It is the final output of the industrial chain. The technical content, quality and market adaptability of the product directly reflect the level of the industrial chain.8) Logistics: It is an important link to connect the upstream and downstream of the industrial chain, including transportation, warehousing, distribution, etc. An efficient logistics system can reduce costs and improve the market competitiveness of products.
    \item[8)] \textbf{Logistics:} It is an important link connecting the upstream and downstream of the industrial chain, including transportation, warehousing, distribution, etc. An efficient logistics system can reduce costs and improve the market competitiveness of products.
    \item[9)] \textbf{Information:} It is an important support for the operation of modern industrial chain, including market information, supply and demand data, logistics information, etc. Effective information transmission and processing can improve the response speed and decision-making efficiency of the industrial chain.
    \item[10)] \textbf{Capital:} It is the material basis for the operation of the industrial chain, involving investment, financing, income distribution and other links. The flow efficiency and allocation efficiency of funds are crucial to the development of the industrial chain.
    \item[11)] \textbf{Talent:} It is the most creative element in the industrial chain, including research and development personnel, management personnel, technical workers, etc. The quality and structure of talents determine the innovation ability and development potential of the industrial chain.
    \item[12)] \textbf{Service:} It is the key to increase the cohesion of enterprise value and extend the service life of products in the industrial chain, which can enhance the added value of products and enhance the competitiveness of enterprises.
\end{itemize}

These key element nodes in the industrial chain is running in a complex dynamical network environment, forming a dynamics with discrete time, space, and state, where spatial interaction and temporal causality are localized. These key element nodes evolve their dynamic properties by adhering to the following physical dynamics rules:

\begin{itemize}
    \item[1)] \textbf{Uncertainty of dynamics:} In the industrial chain, the innovation and development of each element (such as technology, market, competition, policy, etc.) can be seen as following a certain dynamic process, which is also affected by the law of uncertainty, for example, whether a new product can be accepted by the market, and the degree of acceptance, there is great uncertainty. This uncertainty makes the development and change of the industry unpredictable.
    \item[2)] \textbf{Roughness of information dissemination:} In the process of information transmission in the industrial chain, due to the diversification of information sources, information from different sources may have contradictions or conflicts, resulting in the lack of sufficient accuracy of information, some key information may not be collected or recorded at all, information may not be updated in time, limited coverage of data, subjectivity of data interpretation and other problems. As a result, decision makers cannot make decisions based on the latest information and cannot reflect the latest market changes.
    \item[3)] \textbf{Randomness of dynamic behavior:} Each element in the industrial chain is affected by a variety of uncontrollable internal and external factors, such as natural disasters, health crises, political turmoil, etc., resulting in the dynamic behavior or disturbance of each element in the industrial chain having a certain degree of randomness and certainty constantly alternating, and the continuous disturbance, instability, percolation and other phenomena of the overall structure of the industrial chain have formed the vulnerability of the industrial chain.
    \item[4)] \textbf{Confounding of discrete and continuous evolution:} The industrial chain is composed of a number of interconnected links, each including production, distribution, exchange and consumption and other processes, each component changes over time. This evolution is accompanied by the interweave of discrete events and continuous and progressive development. Understanding and describing the discrete and continuous confounding of the evolution of each element of the industrial chain will help enterprises and policy makers to better grasp the development trend of the industrial chain and formulate corresponding strategies and policies to adapt to and guide the healthy development of the industrial chain.
    \item[5)] \textbf{High dimensionality and distribution of state variables:} Each element in the chain has a series of state variables that describe the characteristics and performance of the element at different points in time. The high dimensionality of state variables refers to the fact that these variables constitute a high-dimensional space, while the distribution refers to the distribution of these variables among different elements. Understanding and mastering the high dimension and distribution of each factor state variable in the industrial chain is crucial for the optimization, risk management and strategic planning of the industrial chain. This needs to be achieved through data analysis, model construction and system simulation.
    \item[6)] \textbf{Strong coupling between subsystems:} Each element in the industrial chain is usually composed of multiple subsystems, and there is interdependence and interaction among these subsystems, and the tightness of this mutual relationship is called coupling. Strong coupling means that changes in any subsystem of the industry chain may have a chain reaction, affecting the stability and efficiency of the entire industry chain
\end{itemize}

Given giant chaotic network has \(N\) self-dynamics of node, we define cognitive cellular automaton (CCA) to abstract the dynamic behavior of these key element nodes in the industrial chain. It is formally defined as a quadruple $CA=(C_d ,S ,N ,f)$, which is described as follows:

\begin{itemize}
    \item[1)] \textbf{CCA vector space $\mathbf{C_d}$:} It refers to the finite set of symbolic logic and cognitive operations of the element subject, which are evenly distributed in the discrete first-order or higher-order space, mainly including two kinds of logical operations such as spatial geometry and boundary conditions.The geometric structure includes one-dimensional and multi-dimensional cellular automata. For one-dimensional space, the spatial geometric structure only has a straight line form, but for multi-dimensional cellular automata, the main symbol space structure of elements is arranged into a fully homogeneous sub-network. For two-dimensional space, for example, the grid arrangement can be divided into three forms: triangle, square and hexagon. Boundary conditions can be divided into two forms: boundary-free and cyclic boundary. Boundary-free form means that the symbol of the element body outside the boundary is located in a state of "nothing", while the cyclic boundary form regards the symbol space of the element body as a circle structure. The large-scale CCA in the industrial chain, such as raw materials, information, technology, investment, capital, talent, research and development institutions, are divided into N heterogeneous subgroup vector space, denoted as $\left \langle V_1,d_1 \right \rangle ,\left \langle V,d_1 \right \rangle ,\dots ,\left \langle V_n,d_n \right \rangle $, The thoughtful cellular vector space is defined as $\mathbf{C_d}=V=\left \{ v_1,v_2,\dots ,v_n \right \} (1\le i\le \left | V \right | )$.
    \item[2)] \textbf{CCA state space $S$:} It refers to the finite set of symbolic operating states of each element subject in the industrial chain. In this part, we use negative entropy function to represent the resilience state of each element subject in the industrial chain, which is denoted as $- {\textstyle \sum_{n} }p(n)\log{p(n)} $. If the resilience state of the subject is negative entropy, then the function of the thought cell is in a stable and ordered state; if the resilience state of the subject is positive entropy, then the function of the thought cell is in an unstable and disordered state. The main state of each element of the industrial chain changes at each discrete time step, and the state value of each moment depends on and only depends on the state value of itself and its neighbors at the previous moment.
    \item[3)] \textbf{Adjacency logical relationship $\mathbf{N}$:} It refers to the interaction and adjacent logical relationship between the subjects of each element in the industrial chain in the thought cell space, which is denoted as $\mathbf{N} =(\overrightarrow{v_1} ,\overrightarrow{v_2},\dots ,\overrightarrow{v_n})$.
    \item[4)] \textbf{Behavioral dynamics function $f$:} It refers to the dynamic potential function that determines the evolution of the main state of the element at the next moment according to the current state of each element in the industrial chain and its neighbor status, which is denoted as $f:s_i^{t+1}=f(s_i^t,s_N^t,\gamma )$, where $s_N^t$ is the combination of neighbor states at the moment and $\gamma$ is the control parameter given by the system. In this section, all thought logic evolution rules are summarized as follows:
    \begin{itemize}
        \item[(a)] Stationary rule: Thoughtful cellular starts from any initial state and runs for a certain period of time, the space of thoughtful cellular tends to a spatially stable configuration, which does not change with time, and is denoted as: $\forall s_i,s_n,\gamma \in \mathbf{C_d},v_{(s_i,s_n,\gamma) }\wedge v_{(s_{i+1},s_n,\gamma) }\Rightarrow v_{(s_{i+1},s_n,\gamma) } $.
        \item[(b)] Periodic rules: After running for a certain period of time $t$, thoughtful cellular tend to a series of simple fixed structures Stable Paterns or periodic Perlodical Patterns, denoted as: $\forall s_i,s_n,\gamma \in \mathbf{C_d}, v_{(s_{i},s_n,\gamma) }\Rightarrow v_{(s_{i+1},s_n,\gamma) }$.
        \item[(c)] Chaotic rule: From any initial state, after a certain period of time, thoughtful cellular shows chaotic non-periodic behavior, and the statistical characteristics of the generated structure do not change, which is denoted as: $\forall s_i,s_n,\gamma \in \mathbf{C_d}, v_{(s_{i},s_n,\gamma) }\Rightarrow v_{(s_{i+1},s_{n+1},\gamma) }$.
        \item[(d)] Complex rules: There are complex local structures, or local chaos, some of which propagate constantly.From another perspective, CCA can be regarded as a dynamical system, so a series of concepts such as initial pilot, orbit, fixed point, periodic orbit, and ultimate orbit can be used in the study of cellular automata, which is denoted as: $\forall s_i,s_n,\gamma \in \mathbf{C_d}, v_{(s_{i},s_n,\gamma) }\Rightarrow v_{(s_{i+1},s_n,\gamma_j) }$.
    \end{itemize}
\end{itemize}

In terms of CCA dynamic model modeling, this part of CCA position $\mathbf{V}$ and state $\mathbf{S}$ evolution process at time $t$ will build a dynamic equation of CCA evolution by solving the Hamilton equation, which can realize the transformation of the physical logic symbolic reasoning analysis calculation of the industrial chain into a neural network calculation problem. The dynamic model of CCA is formally described as follows:

\begin{align}
    \frac{dv}{dt} & =  \frac{\partial \mathcal{H } }{\partial s}  ,\frac{ds}{dt}  =  \frac{\partial \mathcal{H} }{\partial v}
\end{align}

\begin{align}
 \mathcal{L}_{HNN}=\left \| \frac{\partial \mathcal{H}_{\theta } }{\partial s}-\frac{\partial v}{\partial t}   \right \| _2 -\left \| \frac{\partial \mathcal{H}_{\theta } }{\partial v}-\frac{\partial s}{\partial t}   \right \| _2
\end{align}

In order to improve the ability of CCA to simulate the evolution mechanism of industrial factors, we defined the observed $v_O$ state and hidden $v_H$ state. The evolutionary dynamic potential function of CCA obtains the evolutionary rules of thought cellular machine through the combination of variational EM reasoning and learning. The objective function is defined as follows:

\begin{align}\label{pvovh}
p(v_O,v_H)= {\textstyle \prod_{(h,r,t)\in O\cap U}} Ber(v_{(h,r,t)}|f(x_h,x_r,x_t))
\end{align}
where $Ber$ represents the Bernoulli distribution, $f(x_h,x_r,x_t)$ computes the probability that the triples are true, and $f(\cdot,\cdot,\cdot)$ represents the scoring algorithm for entity and relation embedding, which varies according to different knowledge graph embedding methods.

It is trained by maximizing the observed log-likelihood value of the indicator variable, i.e., $\log p_w(v_O)$. However, since it is not feasible to integrate all hidden indicator variables $v_H$, this log-likelihood value cannot be optimized directly. Therefore, the evidence lower bound (ELBO) of the log-likelihood function is optimized, and the results are as follows:

\begin{align}\label{logpw}
\log p_w(v_O)\ge E_{q_{\theta }(v_H)}[\log p_w(v_O,v_H)-\log q_w(v_H)]
\end{align}
where $q_{\theta}(v_H)$ is the variational distribution of the hidden variable $v_H$. This equation holds when $q_{\theta}(v_H)$ is equal to the true posterior distribution $p_w(v_O,v_H)$. This lower bound can be efficiently optimized using the variational EM algorithm, which consists of a variational E inference step and an M learning step. The variational E step, known as the inference procedure, fixes $p_w$ and updates $q_{\theta}$ to minimize the KL divergence between $q_{\theta}(v_H)$ and $p_w(v_O,v_H)$. During M-step learning, $q_{\theta}$ is fixed and $p_w$ is updated to maximize the log-likelihood function of all triplets, which is $H_{q_{\theta}({v_H})}[\log p_w(v_O,v_H)]$. Through the learning and reasoning calculation of EM, the CCA group layer of the large model is constructed. The key steps of learning and reasoning of EM are as follows:

In the variational inference calculation step of CCA, this step aims to derive the posterior distribution $p_w(v_O,v_H)$. Since the posterior distribution is difficult to derive accurately, $q_{\theta}(v_H)$ and $p_w(v_O,v_H)$ are used to approximate it. To further improve inference, $q_{\theta}(v_{(h,r,t)})$ is parameterized using amortized inference and knowledge graph embedding model. Formally, $q_{\theta}(v_H)$ is derived as follows:

\begin{align}
q_{\theta }(v_H)= {\textstyle \prod_{(h,r,t)\in H}}q_{\theta }(v_{(h,r,t)})= {\textstyle \prod_{(h,r,t)\in H}}Ber(v_{(h,r,t)}|f(x_h,x_r,x_t))
\end{align}
where the KL divergence between the distribution $q_{\theta}(v_H)$ and the true posterior $p_w(v_O|v_H)$ is obtained by minimizing variation, and the optimal $q_{\theta}(v_H)$ is given by the fixed point condition as follows:

\begin{align}\label{logq}
\log q_{\theta }(v_{(h,r,t)})=E_{q_{\theta }(v_{(h,r,t)})}[\log p_w(v_{(h,r,t)}|v_{MB(h,r,t)})]
\end{align}
where $MB(h,r,t)$ is the Markov blanket of $(h,r,t)$. To update $q_{\theta}$ with $p_w(vO|v_H)$ as the objective, we have the following objective function:

\begin{align}\label{ou}
o_{\theta ,U}= {\textstyle \sum_{(h,r,t)\in O}} E_{p_w}(v_{(h,r,t)}|v_{MB(h,r,t)})[\log q_{\theta }(v_{(h,r,t)}]
\end{align}

While augments the model with instances of observed triples, with the following objective function:

\begin{align}
o_{\theta ,U}= {\textstyle \sum_{(h,r,t)\in O}} \log q_{\theta }(v_{(h,r,t)}=1)
\end{align}
where the derivation of the variational E-step and M-step posterior distributions, the objective function is finally obtained as $O_{\Theta}=O_{\Theta,U}+O_{\Theta,L}$

In the variational learning computation step of CCA, $q_{\theta}$ is corrected, and the weight of the logical rule $w$, which is $E_{q_{\theta}(v_H)}$, is updated by maximizing the log-likelihood function. However, due to the need to deal with partition functions, the pseudo-likelihood function $E_{q_{\theta}(v_H)}[\log p_w (v_{(h,r,t)}|v_{MB(h,r,t)}]$ is directly optimized. Using the gradient descent algorithm to optimize $w$, for each expected conditional distribution $E_{q_{\theta}(v_H)}[\log p_w (v_{(h,r,t)}|v_{MB(h,r,t)}]$, assuming $v_{(h,r,t)}$ interacts with $v_{MB(h,r,t)}$ through a series of rules, the derivative $w_l$ of each rule $l$ is calculated as follows:

\begin{align}\label{wl}
    \nabla _{w_l}E_{q_{\theta}(v_H)}[\log p_w (v_{(h,r,t)}|v_{MB(h,r,t)}]
    \cong y_{(h,r,t)}-(v_{(h,r,t)}=1|v_{MB(h,r,t)})
\end{align}
where the gradient descent algorithm is used to optimize $w$. For each expected conditional distribution $E_{q_{\theta}(v_H)}[\log p_w (v_{(h,r,t)}|v_{MB(h,r,t)}]$ of the potential function of logical evolution dynamics, assuming $V_{(h,r,t)}$ interacts with $v_{(h,r,t)}$ through a series of rules, the derivative $w_l$ of each rule $l$ is calculated as follows:

\begin{align}
\nabla _{w_l}E_{q_{\theta}(v_H)}[\log p_w (v_{(h,r,t)}|v_{MB(h,r,t)}]\cong y_{(h,r,t)}-(v_{(h,r,t)} & = 1|\hat{v} _{MB(h,r,t)})
\end{align}

Through the above modeling, raw materials, markets, technology, policies, innovation, enterprises, products, logistics, information, capital, talents, services and other types of industrial entities and innovative element nodes in the industrial chain are encapsulated into large-scale CCA. As the basic cognitive thought unit of the industrial brain, each CCA has the ability of independent learning and reasoning, and can intelligently represent the operation mechanism of multiple types of industrial entities and innovative element nodes. Therefore, in the CCA encoder layer, we adopt Transformer's Multi-Head Self-Attention coding mechanism to conduct space-time coding for large-scale CCA of raw materials, markets, technologies, policies, innovations, enterprises, products, logistics, information, capital, talents, services, etc. Thus, the thought elements group of the cognitive thought large model is formed. It can obtain the operation mechanism of the industrial chain, industry knowledge and experience from a large amount of data, and automatically adapt to the changing environment, so as to achieve better autonomy.

\subsubsection{Cognitive function network domain layer}

In cognitive function network domain layer, this section defines thought functional space, inspired by the theory of brain cognitive functional area division, which are marked as $\left \langle X_1, d_1\right \rangle ,\left \langle X_2,d_2 \right \rangle ,\dots ,\left \langle X_n,d_n \right \rangle $. Each thought functional space $\left \langle X_i, d_i\right \rangle$ is defined as $A(V,E)$, where $V=\left \{ v_1,v_2,\dots v_i,\dots  \right \} (1\le i\le  | V | )$ represents the set of cognitive cellular automaton in the subspace, and $E$ represents the set of simple complex of the subspace $E=\left \{ e_1,e_2,\dots e_i,\dots  \right \} (1\le i\le  | E| )$. Then each subspace $\left \langle X_i, d_i\right \rangle$ hypergraphs adjacency matrix is defined as follows:

\begin{align}\label{hij}
[h_{i,j}]_{|V|\times |E|}=\begin{cases}
 1 & \text{ if } v\in e \\
 0 & \text{ otherwise }
\end{cases}
\end{align}
where the degree of each node matrix $v$ of the subspace is $d(v)= {\textstyle \sum_{e\in \mathcal{E} }} \omega (e)H(v,e)$, and the degree of each simple complex $e\in \mathcal{E}$ is $\delta (e)= {\textstyle \sum_{v\in V}} H(v,e)$, then the initial form is $X^0=\left \{ x^0_1,x^0_2,\dots ,x^0_N \right \} ,x^0_i \in \mathbb{R}^{C_0} $, and $C_0$ represents the topological eigenspectrum dimension. From macro-scale, mesoscale and micro-scale perspectives, N-heterogeneous hypergraphs\cite{2023HGNN} are dynamically correlated hierarchically and domainically through the multi-head self-attention mechanism, where form thought function collaborative interactive network, which is formally defined as a multi-layer self-attention polarization composition matrix $G(A,O,\mathcal{T})$:

\begin{align}\label{gao}
G(A,O,\mathcal{T})=\begin{bmatrix}
  A^{[1]}\mathcal{T}& O^{[1,2]} & \dots  & O^{[1,M]} \\
 O^{[2,1]} & A^{[2]}\mathcal{T} & \dots  & O^{[2,M]}\\
 \vdots  & \vdots  & \ddots  & \vdots \\
  O^{[M,1]}& O^{[M,2]} & \dots  &A^{[M]}\mathcal{T}
\end{bmatrix}
\end{align}
where $A=\left \{ A^{[1]} ,A^{[2]},\dots,A^{[M]} \right \} $ represents the set of adjacency matrices in the multi-layer network, $A^{[\alpha]}=(V^{[\alpha]},E^{[\alpha]})$ represents the set of CCA group activities in layer $\alpha$, and $E^{[\alpha]}$ represents the set of intra-layer simple complex $\mathcal{T}$ in layer $\alpha$. If CCA node $i$ and CCA node $j$ in layer $\alpha$ have something in common, $a^{[\alpha]}_{ii}=1$, otherwise, $a^{[\alpha]}_{ii}=0$. $O=\left \{ O^{[1,2]},O^{[1,3]},\dots ,O^{[\alpha ,\beta ]} |\alpha \ne \beta \right \}$  indicates the set of adjacency matrices between network layers. $O^{[\alpha,\beta]}=(V^{[\alpha]},V^{[\beta]},E^{[\alpha,\beta]})$, whose functional domain represents $O^{[\alpha,\beta]}_{ii}$ indicates whether there is a dynamic mapping relationship between CCA node $i$ in layer $\alpha$ and CCA node $j$ in layer $\beta$. $V^{[\alpha]}$ and $V^{[\beta]}$ represent the CCA sets of layers $\alpha$ and $\beta$ respectively, and $E^{[\alpha ,\beta ]}=\left \{ (v^{[\alpha ]}_i,v^{[\beta ]}_i)|i,j\in \left \{ 1,2,\dots  \right \};\alpha ,\beta \in \left \{ 1,2,\dots ,M \right \}   \right \} $ represents the set of edges between layers $\alpha$ and $\beta$. The dynamic evolution of the industrial chain is characterized by chaos, criticality, fractal, nonlinearity and sudden change.

The \(N\) gated attention aggregators are set up in a multi-layer multi-domain dynamic network. Through a gated attention aggregator, the dynamic potential energy equation of large-scale CCA is mapped to heterogeneous hypergraph $h_{ij}$ according to different task types. Then, the large-scale CCA state sequence can be regarded as a path, or a process, or a function that changes over time, and is dynamically projected to each layer and each domain in the multi-layer and multi-domain dynamic association network to construct N thought function domains. Each thought function domain consists of one or more CCA and a gated attention aggregator, then the gated attention aggregator $g^k(x)$ is formally defined as:

\begin{align}
    g^k(x)=Softmax(W_{gk}x)
\end{align}
where $W_{gk}\in \mathbb{R}^{n\times d}$ is the trainable memory matrix of CCA, n is the number of thought function, and d is the grid dimension of the input element thought cell space. The gated attention aggregator calculates the weight of each CCA to match the task goal, and can automatically assign cognitive training tasks to several hybrid CCA. By calculating other gated attention aggregators to assign different weights to different attention heads, this provides a possibility to reconstruct the network itself by learning about the network structure.

Through the dynamic combination mechanism of large-scale CCA and multiple gated attention aggregators, large-scale CCA clusters are dynamically divided into different cognitive function domains to form the thought function network layer of the large model, and the cognitive reasoning analysis engine is equipped for the industrial brain. When the cognitive thought system receives the input task data, the gated attention aggregator allocates the input data to multiple thought function domains according to the task type through a simple linear transformation of $Softmax()$ activation function. Then the gated attention aggregator activates one or more CCA clusters to focus on specific target tasks. Thus, a three-layer dynamic heterogeneous hyper network system is formed, which includes functional layers such as thought element group, thought function network and autonomous cognitive interaction. In this architecture, only a few CCA are activated or utilized by input data, while others remain inactive, and the message-passing topology between layers can be dynamically scaled. Thus, the accuracy and efficiency of cognitive reasoning of large models can be improved.

\subsubsection {Cognitive multi-head thought motif layer}

In the autonomous cognitive interaction layer of the LMACT,  the thought motif is designed using Neural ODE, so that automatically generate a variety of human-computer interaction templates. Those templates can help the LMACT to establish the "human in the loop" autonomous interaction mechanism. The interaction motifis $M^*_{i}(K)$ is formally defined as follows: 

\begin{align}
    M^*_{i}(K) &= {\textstyle \prod_{k  = 1}^{K}} \sigma ( {\textstyle \sum_{j\in \Gamma _i}}a^k_{ij}\mathcal{T}^k V_j  )+f(h_t,\theta _t) \label{mi}\\
    a_{ij} &=\frac{e^{LeakyRELU(a([TH_i][TH_j]))}}{ {\textstyle \sum_{k\in \Gamma _i}e^{LeakyRELU(a([TH_i][TH_j]))}} } \label{aij}\\
    \frac{dh(t)}{dt} &=f(h(t),t,\theta)
\end{align}
where $V_j$ and $V^*_i$ represent the simplicial complex self-coding of each important node $i$ at different times, $\Gamma_i$ represents the simplicial complex self-coding of the node $i$ over a binary field, and $K$ represents the number of labels of multi-head attention. If there is an $m-1$ simplicial complex consisting of a chain of $m+1$ significant nodes, the input and output channels of each cognitive interaction module are linearly coupled. Then the topology matching activation function ELU will activate the CCA of the corresponding cognitive functional domain to parameterize the continuous dynamics of the hidden unit, complete the model training by solving the ordinary differential equation, and transmit information to other adjacent cognitive functional domains.

The self-attention mechanism and the interaction topological pairing activation function ELU are introduced in the process of large-scale cognitive interaction module interaction message passing, which is defined as:

\begin{align}\label{fkx}
    f^k(x)= {\textstyle \sum_{i=1}^{n}}g^k(x)_i\cdot f_i(w,a)
\end{align}

in the topological transfer process of each cognitive interaction module, if each gated attention aggregator only selects a CCA group with the highest score, it is organized according to the graph structure in the topological space for relational reasoning. Then each gated attention aggregator divides the input space linearly into N regions, each corresponding to a CCA group. In this way, each CCA group is automatically activated under a given input situation, and the information is transmitted to each CCA group after being selected by a gated attention aggregator, thus building a self-attention cognitive interaction layer of a large model. This allows each CCA group to process inputs according to its design and parameters, while the other CCA groups remain inactive. Thus, the efficiency of large model inference is improved and the calculation cost is reduced.

In order to ensure the accuracy of topological pairing during the dynamic interaction of large-scale cognitive interaction modules, the interactive topological homology group invariance is defined as:

\begin{align}\label{minf}
    arg\min_f \left \{ \mathcal{R}_{emp}(f) +\Omega(f) \right \}
\end{align}
where $\Omega(f)$ is the probability distribution of the optimal initial information propagation at the node of the cognitive interaction model, and $\mathcal{R}_{emp}(f)$ represents the supervisory loss function and the group aggregation function. Then $\Omega(f)$ is formally defined as follows:

\begin{align}\label{omega}
    \Omega(f)=\frac{1}{2} {\textstyle \sum_{e\in \mathcal{E} }}  {\textstyle \sum_{(u,v)\in V }}\frac{\mathcal{T}(e)H(u,e)H(v,e) }{\delta (e)} (\frac{f(u)}{\sqrt{d(u)} }-\frac{f(v)}{\sqrt{d(v)} } )^2
\end{align}
where $\Theta =D^{-1/2}_vHWD^{-1}_eH^TD^{-1/2}_v$ and $\Delta=I-\Theta$, and then the normalized function $\Omega(f)$ reconstructs to $\Omega(f)=f^T\Delta f$, then the positive definite sum $\Delta$ becomes the hypergraph Laplacian matrix. The reconstruction process can be divided into the following three steps: (1) Establishing the likelihood function based on the observed data; (2) The connection probability of topological messages of cognitive interaction modules is obtained by calculating the Expectation-Maximization (EM) likelihood function; (3) Implement an improved two-step reconstruction strategy to significantly improve reconstruction efficiency. The cognitive interaction module interaction structure is obtained, and each cognitive interaction module is represented by the $H()v,e$ boundary matrix, then each cognitive interaction module chain $K= {\textstyle \bigcup_{i\in I}}\sigma _i $,when the vertices of $K$ are formed by all $\sigma_i$, denoted $V(K)$. The $K$ simplicial complex consists of all these $\sigma_i$ connected surfaces, denoted by $E(K)$. In this way, the chain topology $\mathcal{T}(K)$ of each cognitive interaction module is obtained, and then the local aggregation coefficient centered on node can be described by $G(V(K),E(K),\mathcal{T}(K))$ for each thought cognitive interaction circle. The graph attention network is used for interactive edge prediction, and edge prediction is not the ultimate goal we want to achieve. The adjustment is made through the effect of edge prediction. In fact, we can learn much richer information about each node by connecting edges without giving direct information.

\subsection{Human-like autonomous decision-making}
In this section, following Schrodinger's free energy principle \cite{Friston2019free},  we define the "human-in-the-loop" autonomous interactive decision-making model for industrial brain, which consists of four functional modules: situation autonomous awareness, comprehensive situation cognition, co-evolution planning decision-making, and co-evolution topology planning action. The framework integrate four functional modules into the autonomous computing CT-OODA loop \cite{Johnson2023automating}, and employs a multi-head self-attention network to fuse the representations learned from various trajectories dynamically,  Subsequently, it uses a dimension reduction interactive network to project the aggregated representation to a 1-dimensional k-space, providing capabilities of autonomous cognitive planning and adjustment for resilience disaster of industrial chain in a multi-agent dynamic open environment, facilitating accurate prevention of resilience disasters with linear classifiers. The calculation principle of each function module is as follows:

\subsubsection{Running situation observation}

% \newtheorem{step}{Step}

% \begin{step}
%    \textbf{Situation Observation:}
% \end{step}

Through sensors, RFID, bar code scanners and other Internet of Things devices, access to raw materials, markets, technology, policies, innovation, enterprises, products, logistics, information, capital, talent, services and other links in the industrial chain of various types of industrial entities and innovative elements of the real-time data source. The CCA model is used to detect the network dynamic lineage of industrial entities and innovative elements $H(V_t,E_t)$ contained in the snapshot data stream of $k$ frame propagation at $t$ time of the industrial chain. The specific monitoring content is as follows:

 \begin{itemize}
     \item[1)] \textbf{Node spatial-temporal state characteristics $\Theta_{z_1}$}:
     It is used to monitor the local adjacency relationship between industrial entities and innovation element nodes in the system. By measuring the participation coefficient state loss function $\Theta_{z_1}$, we evaluate the entropy change of the spatial-temporal distribution state of nodes in different layers, and obtain the spatial-temporal distribution characteristics of the node community structure of key industrial entities and innovation elements.
     \item[2)] \textbf{Network spatial-temporal clustering coefficient $\Theta_{z_3}$}:
     It is used to monitor the tightness of the local connection structure of industrial entities and innovation element nodes. By solving the aggregation coefficient cross entropy loss $\Theta_{z_3}$, we evaluate the overlap, correlation and centrality of the local connection community structure of industrial entities and innovation element nodes.
     \item[3)] \textbf{Multi-layer network propagation characteristics $\Theta_{\mathcal{P}}$}:
     It is used to monitor the diversity of node topology of large-scale industrial entities and innovative element nodes in multi-layer networks. Through the combination of the maximum coupling strength $q_{max}$ and the effective coupling strength $q^{eff}_c$, the multi-layer network topology robustness loss function $\Theta_{\mathcal{P}}$ is solved, which is used to evaluate the percolation phase change factors of the multi-layer network propagation characteristics.
 \end{itemize}

Assume that in time series $t=T-N+1,\dots,T$, the $k$ frame propagation snapshot data flow matrix $\mathcal{X}_{k\times N\times C_{in}}= \left \{ x_{t1}, x_{t2},\dots ,x_{tk}\right \} $ is collected, where $x_{tk}=\left \{ x_{tk,1},x_{tk,2},\dots ,x_{tk,N} \right \} $, $N$ represents the number of network nodes, $C_{in}$ represents the number of acquisition channels, $n\in N$. Then, the node spatial-temporal state characteristics $\Theta_{z_1}$, network spatial-temporal clustering state $\Theta_{z_3}$ and multi-layer network propagation state $\Theta_{\mathcal{P}}$ are obtained by solving for every $t$ time. Then the state observation equation of industrial entities and innovation elements $H(V_t,E_t)$ is as follows:

\begin{align}
    \mathcal{Z}_1 &= GLU(CNN)\in \mathbb{R} ^{(k-K_t+1)\times N\times C_h} \label{z1} \\
    \mathcal{Z}_2 &= ReLU(CNN(\mathcal{Z}_1))\in \mathbb{R} ^{(k-K_t+1)\times N\times C_h} \label{z2} \\
    \mathcal{Z}_3 &= GLU(CNN(\mathcal{Z}_2))\in \mathbb{R}^{(k_2K_t+2)\times N \times C_{out}} \label{z3} \\
    \mathcal{P} &= Softmax(Dense(CNN(\mathcal{Z}_3)))\in \mathbb{R}^N \label{p}
\end{align}

Through formula \ref{z1}, \ref{z2}, and \ref{z3}, the network essential lineage characteristics such as time stamp, space stamp, node state, degree centrality, clustering coefficient, feature vector centrality, entropy and diffusion in $k$ frame propagation snapshot data stream $\mathcal{X}$ at time $t$ are obtained respectively. The states of temporal characteristics layer $\mathcal{Z}_1$, spatial characteristics layer $\mathcal{Z}_2$ and spatial-temporal clusterin characteristics $\mathcal{Z}_3$ contained in $\mathcal{P}=\left \{ P_1,P_2,\dots,P_N \right \} $ are obtained by formula \ref{p} for gathered. The spatial-temporal clustering coefficient of each output result of $\mathcal{P}$ is normalized by softmax function, and the probability distribution of node density of each key link is generated. The operating state cross-entropy loss function of key elements of each $\mathcal{P}$ output result is as follows:

\begin{align}
\Theta_{\mathcal{Z}_1 }& = arg {\textstyle \min_{\Theta}} - {\textstyle \sum_{i\in V}} y_i\log (\mathcal{Z} _1) \label{thetaz1}
\\\Theta_{\mathcal{Z}_3 }& = arg {\textstyle \min_{\Theta}} - {\textstyle \sum_{i\in V}} y_i\log (\mathcal{Z} _3) \label{thetaz3}
\\\Theta_{\mathcal{P} }&= arg {\textstyle \min_{\Theta}} - {\textstyle \sum_{i\in V}} y_i\log (P_i)+\max_{x_{tj}}q^{eff}_c \label{thetap}
\end{align}

Formula \ref{thetaz1}, \ref{thetaz3} and \ref{thetap} are used to ensure the sensitivity and accuracy of environmental changes, so as to obtain the operation situation of multi-type industrial entities and innovative factors $\Theta_{\mathcal{Z}_1 },\Theta_{\mathcal{Z}_3 }, \Theta_{\mathcal{P} }$ in various links in a specific period of time.

%\begin{step}
 %   \textbf{Risk Orient:}
%\end{step}

\subsubsection{Comprehensive situation cognition}
The critical point disaster entropy of network structure and function disasters of raw materials, markets, technologies, policies, innovations, enterprises, products, logistics, information, capital, talents, services, etc. in the industrial chain is identified from the operation situation of multi-type industrial entities and innovative factors $\Theta_{\mathcal{Z}_1 },\Theta_{\mathcal{Z}_3 }, \Theta_{\mathcal{P} }$ in various links in a specific period of time. By counting the continuous jump and critical index of critical point disaster entropy, the probability distribution region of possible percolation phase transition is identified. The specific process is as follows:

Assuming that in the time series $t=T-N+1,\dots,T$, from the event source region $o_t$ to the continuous step termination region $d_{t+k}$, the potential percolation phase transition event in a specific region is defined as $P_T=(\Theta_{\mathcal{P}},o_t,d_{t+k},T)$. The probability distribution $y_{tk}$ of successive steps, by continuously counting the potential percolation phase transition entropy $\Theta_{\mathcal{P}}$ in a specific region, is expressed as $CA_{\theta}:[E_{T-N+1,\dots,E_T}]\to y_{tk}$, where $N$ represents the length of the time interval and $\mathcal{F}(\Theta_{\mathcal{P}})\to E \in \mathbb{R}^{dim}$ represents the event evolution low-rank matrix. However, in many practical problems, we can not obtain accurate graph structure data in advance, and we are not clear about the interaction structure of the system, and often only the observation data and time series data about the whole system can be collected. In this case, we need an event dynamic heterogeneous graph $G_{P_T}$ that can automatically construct the event phase transition by using the probability distribution $\Theta_{\mathcal{Z}_1 },\Theta_{\mathcal{Z}_3 }, \Theta_{\mathcal{P} }$ of observed data, which is formally defined as follows:

    \begin{align}\label{gpt}
        G_{P_T}=G_{P,T}+G_{o,T}+G_{d,T}
    \end{align}
where $G_{P,T},G_{o,T},G_{d,T} $ represents the heterogeneous graph of the CCA group of the current percolation phase transition event, the CCA group of the percolation phase transition source and the CCA group of the percolation phase transition target respectively, and $+$ represents the CCA connection operator of the event. For the construction of heterogeneous graph $G_{P,T},G_{o,T},G_{d,T} $, see Sec. \ref{LMACT}.

Based on the formulation of the evaluation strategy library for the probability distribution of the possible porous phase transition in $G_{P_T}$, it can be roughly divided into four categories: stationary rules, periodic rules, chaotic rules and complex rules. For details, see the logical Evolution Dynamics section. Quantitative or qualitative methods are used to assess the identified risk factors and analyze the degree and probability of their possible impact on the industrial chain. Assessment tools may include risk matrix, scoring model, impact analysis model, etc., to determine the critical risk areas of key links in the industrial chain such as raw materials, markets, technologies, policies, innovations, enterprises, products, logistics, information, capital, talents, and services. According to the results of the risk assessment model, the high-risk links in the industrial chain are located and the risk sources are analyzed.

%\begin{step}
%   \textbf{Deductive reasoning:}
%\end{step}
\subsubsection{Autonomous decision-making}
According to the evaluation results of risk orient stage, the inductive reasoning and deductive reasoning methods of cognitive diagram of thought large model are used to extract the dynamic graph structure from the multi-stage network disaster propagation process of the risk critical region of a specific key link. Based on the dynamic graph structure, a comprehensive analysis of the multi-stage network disaster propagation process in the risk critical region of key links is carried out, the change trend of risk factors is tracked in real time, the risk development trend is analyzed, the risk assessment and response strategies are generated and adjusted in time, the risk analysis results are converted into decision support information, and the decision makers are helped to formulate network structure reconstruction plans. The specific process is as follows:

In order to extract the dynamic graph structure of risk propagation in a specific region at a specific time and apply the inductive reasoning of a large cognitive thinking model, three different levels of GATs inside h-session, GATs across h-sessions and GATs across subgraphs are presented. The dynamic graph structure is extracted and inductive inference operations are carried out to generate a variety of dynamic interactive graph attention models for risk propagation. The specific working principle is as follows.

In the GATs inside h-session reasoning analysis stage of the event dynamic heterogeneous graph, the working principle of the attention-level inductive reasoning analysis of the event heterogeneous graph is as follows:

    \begin{equation}\label{hhs}
    \begin{aligned}
         h^{(1)}_{h\_s} &= \oplus ^K_{k = 1}\sigma ( {\textstyle \sum_{i \in O_V}}a^k_{h\_s,i}w^k_if_i )
        \\
        a^k_{h\_s,i} &= \frac{\sigma (w^k\cdot [w^k_hh^{(0)}_{h\_s}\oplus w^k_if_i])}{{\textstyle \sum_{j \in O_V}}\sigma (w^k\cdot [w^k_hh^{(0)}_{h\_s}\oplus w^k_if_i])}
    \end{aligned}
    \end{equation}
where $\oplus$ represents the connection operation, $O_V$ represents the set of different types of events in the h-session session phase, and $k$ represents the number of attention tag headers. $h^{(1)}_{h\_s}$ indicates the hidden status of an h-session that passes GAT within an H-session. $h^{(0)}_{h\_s}$ represents the initial state of the event node of h-sessions, and $a^k_{h\_s,i}$ represents the attention weight calculated by the k head between $h\_s$ and event $i$ through the attention mechanism. $w^k$ is the attention matrix in the k head. Due to the heterogeneity of the features of the initial event nodes, $w^k_h,w^k_i$ are used as the transformation matrix of h-sessions, and the initial node features are used as $f_i$ event $i$  (i.e., P, R, F, C) in the k head, and the initial node features are used in the k head.

In the GATs across h-sessions inference analysis stage of the event dynamic heterogeneous graph, for h-sessions sessions in different subgraphs, the working principle of the graph attention cross-layer inductive inference analysis is as follows:

    \begin{equation}\label{h2hst}
        \begin{aligned}
            h^{(2)}_{h\_s_T,\mathcal{P} } &=GAT_p(h^{(1)}_{h\_s_{T-t},\mathcal{P} }),\forall t,0\le t\le N_p\\
            h^{(2)}_{h\_s_T,O} &=GAT_o(h^{(1)}_{h\_s_{T-t},O}),\forall t,0\le t\le N_o\\
            h^{(2)}_{h\_s_T,d} &=GAT_d(h^{(1)}_{h\_s_{T-t},d}),\forall t,0\le t\le N_d
        \end{aligned}
    \end{equation}
where $N_p,N_o,N_d$ represents the number of different timestamps of the event in the current state, initial state and end state respectively. At the same time, $h^{(2)}_{h\_s_T,\mathcal{P} },h^{(2)}_{h\_s_T,O},h^{(2)}_{h\_s_T,d}$ represents the hidden state of the current event, initial event and end event of h-sessions at time $T-t$. As shown in the equation, different GAT models are trained for different subgraphs. The special subscript t starts from 0, so self-attention is also used in the GAT model across h-sessions. We express the attention mechanism of the $GAT_P$ graph model of the current event as formula \ref{hhs}, while the attention of other subgraphs $(GAT_o,GAT_d)$ is calculated in a similar way.

    \begin{equation}\label{h2hstp}
    \begin{aligned}
         h^{(2)}_{h\_s_T,\mathcal{P} } &= \oplus ^{K_{\mathcal{P} }}_{k = 1} {\textstyle \sum_{t=0}^{N_{\mathcal{P} }}}a^k_{t,T}w^{\mathcal{P},k }_lh^{(1)}_{h\_s_{T-t},\mathcal{P} }
        \\
        a^k_{t,T} &= \frac{\sigma (w^{\mathcal{P},k }\cdot [w^l_ph^{(1)}_{h\_s_T,\mathcal{P} }\oplus w^p_lh^{(1)}_{h\_s_{T-t},\mathcal{P}}])}{{\textstyle \sum_{j=0}^{N_{\mathcal{P} }}}\sigma (w^{\mathcal{P},k }\cdot [w^l_ph^{(1)}_{h\_s_T,\mathcal{P} }\oplus w^p_lh^{(1)}_{h\_s_{T-t},\mathcal{P}}])}
    \end{aligned}
    \end{equation}
where $h^{(2)}_{h\_s_T,\mathcal{P} }$ is used to aggregate all adjacent simplicial complex of a particular event subgraph, and where $w^{\mathcal{P},k}$ represents the linear transformation matrix before attention, $w^{\mathcal{P},k}_l$ represents the attention matrix of the k head in the subgraph.

In the inference analysis stage of GATs across subgraphs of dynamic heterogeneous graphs of events, the inductive inference analysis of global attention for event disaster propagation works as follows:

    \begin{align}\label{hgpt}
        h^g_{P_T}=h^{(3)}_{h\_s_T}=GAT_g(h^{(2)}_{h\_s_T,p},h^{(2)}_{h\_s_T,o},h^{(2)}_{h\_s_T,d})
    \end{align}

where $h^{(3)}_{h\_s_T}$ is the final event for h-sessions at time $T$, and the final event embedding will be updated accordingly after GAT cross-subgraph processing. $GAT_g$ represents the aggregation information function at the fully connected layer of heterogeneous subgraphs. Specifically, the working mechanism of global attention transformation for current events is designed as follows:

    \begin{equation}\label{h3hst}
    \begin{aligned}
         h^{(3)}_{h\_s_T } &= \oplus ^{K_{g}}_{k = 1} {\textstyle \sum_{i\in O_G}^{N_{\mathcal{P} }}}a^k_{i,T}w^{g,k }_lh^{(2)}_{h\_s_{T},t }
        \\
        a^k_{i,T} &= \frac{\sigma (w^{g,k }\cdot [w^{g,k}_lh^{(1)}_{h\_s_T,\mathcal{P} }\oplus w^{g,k}_lh^{(2)}_{h\_s_{T},t}])}{{\textstyle \sum_{i\in O_G}}\sigma (w^{g,k }\cdot [w^{g,k}_lh^{(1)}_{h\_s_T,\mathcal{P} }\oplus w^{g,k}_lh^{(2)}_{h\_s_{T},t}])}
    \end{aligned}
    \end{equation}
where $O_G$ represents different types of heterogeneous graphs, $K_g$ represents the amount of global multi-headed attention, $w^{g,k}$ represents the global attention matrix of multi-headed attention, and $w^{g,k}_l$ refers to the transformation matrix in global attention. Through three different levels of attention, such as GATs inside h-session, GATs across h-sessions and GATs across subgraphs, logical induction reasoning operations are used to obtain the risk propagation law of multi-stage network topology in the risk critical region of specific key links.

The risk analysis results step at a specific time $T$, using the deductive reasoning method of cognitive thinking large model, the system generates a comprehensive research and judgment GRU sequence model of industrial chain network structure reconstruction, and generates a formal description of network structure reconstruction scheme as follows:

    \begin{equation}\label{zt}
        \begin{aligned}
            Z_T  & = \sigma (w_z\cdot E_T\oplus U_Z\cdot h_{T-1})\\
            r_T  & = \sigma (w_r\cdot E_T\oplus U_r\cdot h_{T-1})\\
            \acute{h} _T  & = \tanh (w\cdot E_T \oplus U(r_T\circ h_{T-1} ))\\
            h_T &= (1-Z_T)\circ h_{T-1}+Z_T\circ \acute{h} _T
        \end{aligned}
    \end{equation}
where $Z_T$ represents $E_T$ abstractly synthesized by the global attention mechanism at time $T$, $h^g_{P_T}$ is obtained by \ref{h2hstp}, $\oplus$ represents the connection operation, $\circ$ represents the Hadamard product operation, $h_T \in \mathbb{R}^{dim}$ is the output hidden layer state at time $T$, and $w$ and $U$ are parameters that need to be learned.

At a specific time $T$, system entropy is used to convert the comprehensive research results of industrial chain network structure reconstruction into a decision support minimization loss function, which is formally described as follows:

    \begin{align}\label{l}
        L=- {\textstyle \sum_{u\in U}}  {\textstyle \sum_{t=1}^{T}} y^E_T\log (\sigma (\theta .h^E_T))-(1-h^E_T)\log (1-\sigma (\theta .h^E_T))+\lambda \left \| \theta  \right \|_2
    \end{align}
where $y^u_T$ represents the label of user $u$ at time $T$, $h^u_T$ represents the embedding of event $u$ at time $T$, $\sigma$ represents the sigmoid function, $\theta$ represents the parameter to be learned, and $\lambda \left \| \theta  \right \|_2 $ is the L2 regularization. In our framework, the conversion of risk evaluation results into decision support minimization loss functions is optimized using gradient descent method. Real-time tracking of the change trend of risk factors, research and judgment of risk development trends, generate timely adjustment of risk assessment and response strategies, real-time risk research and judgment results into decision support schemes, and help decision makers to formulate network structure reconstruction schemes.

%\begin{step}
%    \textbf{Executive Planning:}
%\end{step}

\subsection{Human-like autonomous planning action}

The section explore the reverse direction by using a multi-agent logic of knowing how to do know-how-based planning via model checking and theorem First-Order Satisfiability\cite{li2024knowing} checking. Based on our human-like autonomous decision-making, we define human-like autonomous cognitive planning module, which include two new classes of related planning problems: higher-order epistemic planning and meta-level epistemic planning. two key processes are as follows: 

\subsubsection{Node-level epistemic planning}
The industrial chain robustness is to establish on interconnected nodes and weighted links. Node-level epistemic planning is undoubtedly the most direct and intuitive analytical framework for dealing with network robustness problems, and it is crucial for understanding the large-scale node`s robustness\cite{Artime2024robustness} ,when the face of external disturbances and internal failures. At the same time, considering that large-scale node`s dynamic percolation, such as deterministic network and stochastic network, the modeling of percolation transition depends on the expression of phase transition phenomena with the characteristics of degree value connectivity and coupling dynamics. 
A distinct feature of modal logic is that it brings notions of the meta-language to the node`s dynamic. This is not merely formalizing the existing meta-language concepts more precisely since the object language can open new possibilities. Specifically, the universal equation of percolation in interconnected nodes involves the optimization calculation of the maximum connected piece and critical point estimation. The general equation of percolation in node-level epistemic planning is as follows:

    \begin{equation}\label{sst}
        \begin{aligned}
            S^{st}_i &=x_ig_i(x_i)\\
            x_i &=\phi _i {\textstyle \prod_{j=1}^{k}} [q_{ji}y_{ji}g_i(x_i)-q_{ji}+1]\\
            y_{ji} &=\frac{x_i}{q_{ji}y_{ji}g_i(x_i)-q_{ji}+1}
        \end{aligned}
    \end{equation}
where $x_i$ represents the connectivity of the network degree value, and $g_i(x_i)$ represents the dynamic potential function of the network coupling. $q_{ji}$ represents the dependence between the nodes of layer $i$ and layer $j$ of the network, and $y_{ji}$ represents the optimal calculation method for estimating the critical point of connectivity of the network degree value. $k$ refers to the number of layers connected through dependencies. These parameters have a significant effect on the robustness of the interdependent system. The most important path of optimal regulation is the precise calculation of percolation phase transitions such as degree value connectivity and coupling dynamics that affect the Robustness of the industrial chain network in accordance with \ref{sst}, providing a wealth of maximum connected segments and critical point estimation schemes. 

From simply absorbing and digesting small perturbations of specific types without causing large scale effects, to amplifying small initial damages until they extend to affect the entire dynamic system. With the change of network topological parameters, the maximum connected slice undergoes a discontinuous phase transition at the end of the cascade process. This mutation leads to a sudden breakdown of the system, which becomes more difficult to predict as the number of fully coupled subsystems of deterministic and random networks increases. This can be mitigated by reducing the level of interlayer dependence and, if reduced enough, can even restore the continuous disintegration process. More importantly, the analysis of formula \ref{sst} provides insights into how to improve the robustness of codependent coupled systems. Three main approaches have been identified: increasing the proportion of autonomous nodes, especially those with high numbers; Design dependencies between nodes of the same degree; And a special focus on protecting high-value nodes from failures and attacks. In the context of brain networks, developing strategies to prevent and cope with the breakdown of network systems, if topological correlations are taken into account, can improve the robustness of the network to the cascade effects induced by interdependence.

\subsubsection{Topology-level epistemic planning}
The topology-level resilience of the industrial chain is to establish the co-evolution model of network behavior and the optimization method. Considering that the industrial chain in the real world is composed of many nodes topology through complex weighted directed interaction relations, and is controlled by a large number of topology. Therefore, their topology states not only be described by one-dimensional equations, but also need to be coupled nonlinear equations to capture the interaction relationship via underlying network topology. Therefore, the obtained topology-level resilience function is a multidimensional manifold on the complex parameter space of the system. 
An exciting feature of such planning is that you do not need to know what actions are available to others; you just need to plan at the topology-level resilience by making use of others’ knowledge-how, without going into the details about actions. This is what we mean by topology-level epistemic planning: planning without a given model but by formal reasoning based on formulas. Such planning based on who knows how to do what is at the core of the arts of leadership
and management in general. The universal equation of topology-level epistemic planning is as follows:

    \begin{align}\label{dxi}
        \frac{dx_i}{dt}=f(x_i(t))+ {\textstyle \sum_{j=1}^{N}}  A_{ij}H(x_i,x_j)
    \end{align}
where $f(x_i(t))$ describes the self-dynamic dynamic potential function of the active node $x_i(t)$ system at time $t$ of the industrial chain, $A_{ij}H(x_i,x_j)$ reflects the dynamic rules of the interaction between the element $i$ and the neighbor element $j$, and the weighted adjacency matrix $A_{ij}$ captures the interaction rate between all pairs of components. The resilience of multidimensional systems can be captured by calculating the stable fixed points of formula \ref{dxi}, thus building the basic framework of network resilience analysis. In addition, there may be different forms of perturbation that cause the adjacency matrix to change, such as node/link removal, weight reduction. 

This means that the resilience of a multidimensional system depends on the network topology and the form of perturbation that occurs. The state of the dynamical system may change as external conditions change, and the response is usually obvious. Then the optimization regulation of industry chain resilience follows: when the external conditions gradually change over time, the state of some high-resilience systems may show a linear response. The more common and complex response relationships are nonlinear, with continuous and transitional phase transitions "from top to bottom" as the control parameters increase. In both linear and continuous nonlinear cases, there is only one equilibrium state. When the control parameters are restored to the original level, the system can also be restored to the previous state. Some systems may be "unresponsive" within a certain range of conditions, but suddenly respond when the control parameter approaches a certain threshold, exhibiting unexpected state transitions between two independent states. The phase transition is a single state in all environments (although the state may change at the threshold). The existence of alternating stable states means that in the same environment, at least within a certain range, two different stable states can occur, and hysteresis phase transitions often occur. In the case of alternating stable states, the transition between states is characterized by a strong qualitative change, while the phase transition is characterized by a simple quantitative change.

\section{Experiments and discussions}\label{4}
To evaluate the performance of the proposed industrial brain framework, two types of experiments are conducted. The first one experiments are designed for evaluating the parameters and performance of cognitive diagram of thought. The last one is designed for evaluating the performance of autonomous decision-making. In the following, we introduce experimental settings, results, comparisons, and discussions, respectively.

\subsection{Experimental Settings}

\subsubsection{Scenario configuration}
In this experiment scenario configuration, we select realcar parts industry chain of taizhou city in china, as this paper`s experimental validation environment, which has 100 collaborative enterprises\cite{Spieske2021improving} involved in the sales, research and development, production, material supply, after-sales service,.etc. The industrial chain is usually composed of upstream and downstream enterprise value chain, raw material procurement and supply chain, research and development and production chain, logistics and distribution chain, sales and after-sales service chain, etc. These key elements play the following roles in the experimental environment:
\begin{itemize}
    \item[1)] Upstream and downstream enterprise value chain: The chain contains 99 parts enterprises need to work closely with 1 vehicle manufacturer order transactions, raw material procurement and product delivery and other information transfer. It is important to ensure the supply and demand balance of upstream and downstream enterprises in the industrial chain.
    \item[2)] Raw material procurement and supply chain: The chain contains 10,000 different batches of auto parts products, including engines, chassis, body, electronic and electrical systems and other production needs of various metal, plastic, rubber, glass and other raw materials, raw material quality and cost, raw material supply delivery cycle and default and other key elements of information transmission. It is important to reduce the operating cost of the industrial chain.
    \item[3)] Research and development and production chain: The chain includes casting, forging, stamping, welding, painting, assembly and other process research and development, 10,000 different batch production planning and scheduling, 3,000 process monitoring and 400 equipment OEE evaluation and other key elements of information transmission. Through a strict quality management system, we ensure that the quality of parts meets the standards and requirements of automobile manufacturers. It is important to ensure the quality and production efficiency of components in the industrial chain.
    \item[4)] Logistics and distribution chain: The chain mainly involves the logistics distribution, delivery and repair of 10,000 different batches of parts and other key links of information transmission, to ensure that parts can be timely and accurate delivery to the car manufacturer or repair shop. It is important to improve the operation efficiency of the industrial chain.
    \item[5)] Sales and after-sales service chain: The chain involves auto parts manufacturing enterprises to sell vehicles to consumers through a sales network, while providing after-sales services, such as repair, maintenance, spare parts supply, etc. The information transmission of sales and after-sales service links is crucial to maintain the high-quality development of the industrial chain.
\end{itemize}

\subsubsection{Datasets}
In the experimental environment, we select deterministic real-world network datasets from car auto parts industry chain, and random noise network datasets from domain experts randomly adding in this experiment. The deterministic real-world network datasets is composed of upstream and downstream enterprise value chain DS1, raw material procurement and supply chain DS2, research and development and production chain DS3, logistics and distribution chain DS4, sales and after-sales service chain DS5). The characteristics of real-world network datasets is shown in Table \ref{rsd}. 
The random fault synthesis data flow come from supply and demand balance drift RD1, order and capacity drift RD2, planning and execution drift RD3, logistics and inventory drift RD4. The characteristics of random fault synthesis data is shown in Table \ref{cdd}. Those data are collected through customer relationship management (CRM) systems, procurement systems, product lifecycle management (PLM) systems, quality management systems (QMS), logistics management systems (TMS), warehouse management systems (WMS), supply chain management software (SCMS), supply chain management (SCM) systems, enterprise resource planning (ERP) systems and other tools. The detail data characteristics are as follows:
\begin{table}[!hbtp]
	\caption{Real scenario data stream feature}\label{rsd}
	\centering
	\setlength{\tabcolsep}{6.5mm}
	
	\begin{tabular}{c|cccc}
		\toprule
		Data Flow  & Sample size & Feature size & Data distribution & Outliers(\%) \\ % Table header row
		\midrule
		DS1 & 450000000 & 96 & Repeatability & 0\\
		DS2 & 450000000 & 96 & Repeatability & 15\\
		DS3 & 450000000 & 96 & Incremental type& 15\\
		DS4 & 450000000 & 96 & Mutant type & 15\\
		DS5 & 450000000 & 96 & Gradual type & 15\\
		\bottomrule
	\end{tabular}
\end{table}
\begin{itemize}
    \item[1)] Upstream and downstream enterprise value chain DS1: a data set covering the whole process of enterprises from raw material procurement, production, sales to after-sales service, including: Upstream and downstream enterprise information, product quantity, price, delivery time, inventory level, storage cost, product turnover, market demand, after-sales service, new product development, patent application, cost, revenue, supply chain cooperation, market competition pattern and partner evaluation, etc.
    \item[2)] Raw material procurement and Supply chain DS2: a data collection covering the entire process of raw material procurement, inventory, logistics and market fluctuations, including: basic information such as supplier name, address, contact information, material name, specification, model, unit, price, order number, order date, order status (such as: ordered, shipped, completed, etc.), order amount, inventory quantity, inventory location, inventory status (such as: available, reserved, consumed, etc.), logistics company name, logistics order number, delivery date, arrival date, payment method, payment amount, payment date, price change date, reason for change, change range, supplier rating, evaluation date, evaluation content, etc.
    \item[3)] R\&D and production chain DS3: usually involves the whole process from product concept design to final product production, including: R\&D data (R\&D project number, R\&D project name, R\&D goal, R\&D team configuration, R\&D budget, patents, technical papers, etc.); production stage information (production order number, production batch, production date, production quantity, production equipment information, production schedule, quality control data, defective product records); cost and financial information (cost accounting data, financial statements and cost control analysis).
    \item[4)] Logistics and distribution chain DS4: usually involves a series of data related to logistics and distribution process, which helps enterprises and organizations to monitor and manage logistics activities in the supply chain, including: order management information, warehouse management information, transportation management information, distribution center information, packaging and sorting information, quality control and security inspection information, cost and cost data.
    \item[5)] Sales and after-sales service chain DS5: involves a series of links from product sales to after-sales service, and the content of the data set include: sales information, inventory management information, supply chain information, after-sales service information, quality control and after-sales service information, cost and cost data, market and demand information, and legal and compliance information.
\end{itemize}
\begin{table}[!hbtp]
	\caption{Concept drift data stream feature}\label{cdd}
	\centering
	\setlength{\tabcolsep}{6.5mm}
	
	\begin{tabular}{c|cccc}
		\toprule
		Data Flow  & Sample size & Concept drift & Data distribution & Width of drift\\ % Table header row
		\midrule
		RS1 & 450000000 & 96 & Mutant type & 20\\
		RS2 & 450000000 & 96 & Repeatability & 15\\
		RS3 & 450000000 & 96 & Incremental type& 10000\\
		RS4 & 450000000 & 96 & Gradual type & 10000\\
		\bottomrule
	\end{tabular}
\end{table}

The required random disturbance is implemented by four types of random fault synthesis data streams RD1~RD4, which are supply and demand, order and capacity, planning and execution, logistics and inventory, etc., which are non-equilibrium data streams with variable imbalance ratio of network topology, and the class imbalance disturbance ratio changes at the 200000-th sample. 5\% disturbance outliers and 100 concept drift disturbances were added to data streams RD1-RD4, among which the 1st and 3rd were incremental concept drift and the 20-th were mutant concept drift. 10 concept drifts were added to data streams RD1-RD4, which were mutant, repetitive, incremental and gradual concept drift respectively. The network topology data stream generator RandomRBF\cite{Huang2005extreme} and the network topology synthesis tool\cite{Seiculescu2010sunfloor}, including ConceptDriftStream construct mutant and incremental concept drift data streams. The specific steps are as follows: Firstly, RandomRBF is used to cluster the topological datum of multiple subnets, and some clusters are selected from the clustering results as data streams with specific distribution (when the number of network topology clusters is different, the distribution of selected network topology data streams is also different, and they can be used as data streams before and after concept drift). Then the ConceptDriftStream is used to link the data flows distributed in different network topologies. Based on the given network topology concept drift width, the last sample of the previous part of the data flow is extended to both sides, and the given disturbance function is used to simulate the concept drift for this part of the data. The type and amount of network topological drift is unknown.

\subsubsection{Experimental training environment}
\begin{figure}[!htbp]
	\centering
	\includegraphics[width=16cm]{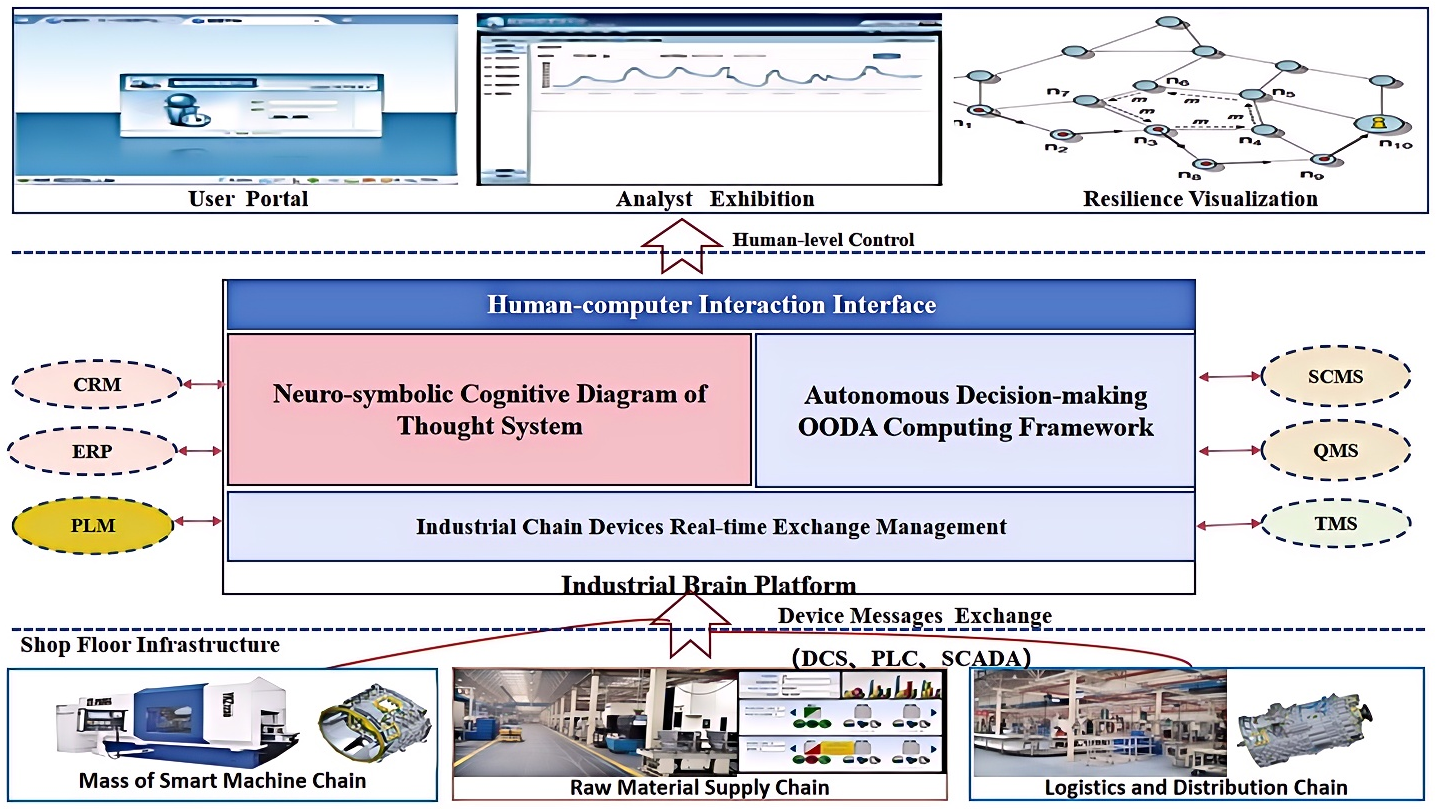}
	\caption{The prototype system of industrial brain. The experimental training run on an iterative three-step closed loop: receiving sensory data from the car auto parts industry chain environment, performing resilience cognitive modeling on historical prior data and decision-making on online measured data, by performing actions that affect the environment. This system is  use to deploy and verify performance of resilience inference and autonomous decision-making on the car parts industry chain.}\label{fig9}
\end{figure}

The experimental training environment in this paper is implemented by the cloud computing environment of Chinese Academy of Sciences. The hardware environment is equipped with 10 NVIDIA A100 GPU processors, 15 processors are Xeon Platinum 8500, 3.90ghz, 128 GB memory capacity, and 200GB data storage capacity. Using PyTorch 2.0\cite{Wu2023pytorch}, Apache Spark2.8\cite{Meng2016mllib}, DeepLogic\cite{Duan2022deeplogic}, Qwen-7B\cite{Pan2024dense} and other toolkits, we developed a prototype system of industrial brain. The prototype system is deployed in the ecological environment of the passenger cars industry chain for verification performance. The specific experimental environment is showed in Figure \ref{fig9}.

\subsection{Resilience inference on the car parts industry chain}

This experiment selected real network data flows (including five categories: upstream and downstream enterprise value chain DS1, raw material procurement and supply chain DS2, R\&D and production chain DS3, logistics and distribution chain DS4, and sales and after-sales service chain DS5). From the four aspects of model recognition accuracy (Accuracy), model inference recall (Recall), model training calculation efficiency (Efficiency), model adaptability (Adaptability), and model interpretability (Interpretability), we compare the performance of the cognitive diagram of thought large model (CDTLM) proposed in this paper with the current mainstream thought methods, such as GoT, CoT and Auto-CoT built based on LLM. 

\subsubsection{Baselines}
In terms of the baseline setting of resilience inference experiments, our proposed CDTLM is compared with the current three groups of mainstream algorithms: (1) LLM model-based Mind map (GoT); (2) Enhanced LLM based on prompt engineering optimization (Auto-CoT); (3) Comparison of enhanced LLM  based on process optimization (SC). To ensure a fair comparison, the three major sets of algorithms chose between five classes of real network data streams and four classes of random fault synthesis data streams, all of which used the same intent-enhancing or voting mechanism. Comparison and analysis were made on their evaluation indicators such as Accuracy, F1 Score, Generalization Ability and Resource Consumption. The current mainstream algorithm is as follows:

\begin{itemize}
	\item[1)] GoT\cite{Besta2024graph}: It is able to model the information generated by the LLM into an arbitrary graph structure where the units of information (" LLM ideas ") are nodes and the edges correspond to the dependencies between these nodes.
	\item[2)] CoT\cite{Wei2022chain}: It is a method for interpreting and understanding the reasoning processes of large language models such as GPT-3. The thought chain makes the decision-making process of the model more transparent and understandable by decoding the internal representation of the model into a series of logical steps.
	\item[3)] Auto-CoT\cite{Aguilera2022particular} : It is an automatic Chain of Thought (CoT) prompt method. This approach involves sampling diverse questions and generating inference chains to build examples.
\end{itemize}

\subsubsection{Evaluation metrics}
The evaluation metrics of resilience inference experiments are specifically defined as follows:
\begin{itemize}
    \item[1)] Model recognition accuracy (Accuracy): It is used to measure the accuracy of the four mental models in this experiment that can correctly identify basic network units such as industrial chain element nodes and links from real network data flow test samples. The usual calculation method is: the proportion of the number of accurately identified industrial chain element nodes and logical structures to all test samples.
    \item[2)] Model inference recall (Recall): It is used to measure the accuracy of the four thought models in this experiment that can classify and retrieve industrial chain theme network communities from real network data flow test samples. The usual calculation method is: the proportion of the number of correctly classified industrial chain theme network communities among all test samples. The level of recall rate is directly related to the degree to which the system or model satisfies user requests.
    \item[3)] Model computing efficiency (Efficiency): It is used to measure the consumption of resources such as CPU, GPU, and memory during the process of building a mental model based on LLM. The calculation method is usually based on the throughput of resource scheduling. The computational efficiency of model construction is determined by solving the average execution resource overhead of completing model pre-training and interactive inference tasks N times at a specific time interval of t.
    \item[4)] Model adaptability (Adaptability): Whether the model can apply the knowledge learned on a specific data set to a new task data set and still maintain stable performance. The combination is evaluated through model accuracy and recall, whether the model can easily adapt to new tasks or fields, and whether it can quickly adapt to new environments or tasks with a small number of samples.
    \item[5)] Model Interpretability (Interpretability): It refers to comparing the thought model of this article with the ROC curve of the causal relationship analysis results of failures in certain links by experts in a specific field to evaluate whether the model is interpretable.
\end{itemize}

\subsubsection{Experimental results and discussions}
Experimental results on all five datasets are provided in
Tables 3, 4,5,and 6. From these results we can have the following
observations:
\begin{itemize}
	\item[1)] In these results from Table \ref{acc} and Table \ref{rec}, our proposed CDTLM adopts the hierarchical network dynamic message mechanism mode combining CCA and gated network, which makes the model recognition accuracy and target retrieval classification recall rate coefficients on the 5 real data streams highest. In particular, the advantages are more obvious on DS3 and DS5, which have a large number of categories in the industrial chain community structure aggregation system. The accuracy coefficients identified on DS3 and DS5 data streams are 10.14\% and 13.30\% higher than the latest GoT algorithm, respectively. The classification recall rate of model retrieval on DS3 and DS5 data streams is 4.50\% and 5.46\% higher than the latest GoT algorithm, respectively. On DS1 data stream with frequent random disturbance, the recognition accuracy coefficient of our LMACT algorithm is 4.06\% higher than that of GoT algorithm, and the retrieval classification recall coefficient of our proposed CDTLM algorithm is 4.06\% higher than that of GoT algorithm. Through multi-dimensional parameter analysis and comparison, it is further shown that our proposed CDTLM algorithm can greatly improve the scale, accuracy and efficiency of the model to deal with complex problems by introducing intelligent experts and sparse network integration.
	\item[2)] In these results from Table \ref{effi}, the Efficiency system value of our proposed CDTLM is superior to other comparison algorithms on all data streams, and the advantages of training convergence and inference convergence efficiency on two data streams with a large number of categories (DS3 and DS5) are more obvious. On DS3 data stream, Efficiency value is 4.08\% higher than GoT, and on DS5 data stream, Efficiency value is 5.56\% higher than GoT.
	\item[3)] In these results from Table \ref{adapt}, the parameter invocation of our proposed CDTLM large model adopts the principle of sparse dropout activation based on the mixed expert model MoE, and chooses to activate a certain number of cellular automata models according to the specific requirements of the task to complete this task, while dropout performs random deactivation of neurons in the neural network. During each training, only certain parameters are retained, which not only makes the network have sparse characteristics, reduces the parameter pressure of the whole network, but also reduces the probability of overfitting of the model and improves the inference generalization ability of the model. The availability coefficient of our proposed CDTLM parameter invocation is superior to other comparison algorithms on all data streams, and the accuracy coefficient and applicability coefficient of interactive inference on two data streams (DS3 and DS5) with a large number of categories are more obvious. On the DS3 data flow, Adaptability is about 2.05\% higher than GoT, respectively, on the DS5 data flow, Adaptability is about 3.02\% higher than GoT.
\end{itemize}

\begin{table}[!hbtp]
\caption{\textit{Accuracy} value of seven alorithms (\%)}\label{acc}
\centering
\setlength{\tabcolsep}{6.5mm}

\begin{tabular}{c|cccc}
\toprule
 Data Flow  &  GoT & Auto-CoT & CoT & \textbf{Our CDTLM}\\ % Table header row
 \midrule
 DS1 & 89.01 $\pm$ 0.21 & 87.03 $\pm$ 0.15 & 86.28 $\pm$ 0.28 & \textbf{93.07 $\pm$ 0.14} \\
 DS2 & 93.31 $\pm$ 0.14 & 89.27 $\pm$ 0.49 & 91.57 $\pm$ 0.36 & \textbf{94.64 $\pm$ 0.26} \\
 DS3 & 79.44 $\pm$ 0.34 & 73.23 $\pm$ 0.56 & 71.44 $\pm$ 0.61 & \textbf{89.58 $\pm$ 0.16} \\
 DS4 & 89.61 $\pm$ 0.28 & 90.34 $\pm$ 0.42 & 89.57 $\pm$ 0.72 & \textbf{91.44 $\pm$ 0.47} \\
 DS5 & 73.34 $\pm$ 0.37 & 74.57 $\pm$ 0.50 & 89.57 $\pm$ 0.73 & \textbf{86.64 $\pm$ 0.18}\\
 \bottomrule
\end{tabular}
\end{table}

\begin{table}[!b]
\caption{\textit{Recall} value of seven alorithms (\%)}\label{rec}
\centering
\setlength{\tabcolsep}{6.5mm}

\begin{tabular}{c|cccc}
\hline
 Data Flow  &  GoT & Auto-CoT & CoT & \textbf{Our CDTLM}\\ % Table header row
 \hline
 DS1 & 92.01 $\pm$ 0.33 & 92.03 $\pm$ 0.27 & 87.28 $\pm$ 0.34 & \textbf{95.57 $\pm$ 0.24} \\
 DS2 & 93.15 $\pm$ 0.52 & 91.13 $\pm$ 0.16 & 88.27 $\pm$ 0.44 & \textbf{94.17 $\pm$ 0.14} \\
 DS3 & 84.24 $\pm$ 0.34 & 83.33 $\pm$ 0.23 & 79.34 $\pm$ 0.45 & \textbf{88.74 $\pm$ 0.23} \\
 DS4 & 90.81 $\pm$ 0.16 & 89.46 $\pm$ 0.44 & 86.23 $\pm$ 0.46 & \textbf{92.24 $\pm$ 0.14} \\
 DS5 & 83.32 $\pm$ 0.17 & 78.74 $\pm$ 0.16 & 76.67 $\pm$ 0.47 & \textbf{88.78 $\pm$ 0.47} \\
 \hline
\end{tabular}
\end{table}

\begin{table}[!htbp]
\caption{\textit{Efficiency} value of seven alorithms (\%)}\label{effi}
\centering
\setlength{\tabcolsep}{6.5mm}

\begin{tabular}{c|cccc}
\toprule
 Data Flow  &  GoT & Auto-CoT & CoT & \textbf{Our CDTLM}\\ % Table header row
 \midrule
 DS1 & 89.13 & 87.26 & 85.16 & \textbf{92.17} \\
 DS2 & 91.33 & 89.41 & 86.13 & \textbf{93.24} \\
 DS3 & 86.24 & 82.33 & 79.33 & \textbf{90.32} \\
 DS4 & 90.41 & 90.22 & 88.43 & \textbf{93.78} \\
 DS5 & 85.52 & 83.78 & 81.33 & \textbf{91.08} \\
 \bottomrule
\end{tabular}
\end{table}

\begin{table}[!htbp]
    \centering
    \caption{\textit{Adaptability} value of seven alorithms (\%)}
    \label{adapt}
    \setlength{\tabcolsep}{3.5mm}
    \begin{tabular}{c|cccccccc}
    \hline
     \multirow{2}*{Data Flow}   &  \multicolumn{2}{c}{GoT} & \multicolumn{2}{c}{Auto-CoT} & \multicolumn{2}{c}{CoT} & \multicolumn{2}{c}{\textbf{Our CDTLM}}\\
    \cmidrule(lr){2-3}\cmidrule(lr){4-5}\cmidrule(lr){6-7}\cmidrule(lr){8-9}
     & 1-shot & 5-shot & 1-shot & 5-shot & 1-shot & 5-shot & 1-shot & 5-shot \\
     \hline
     DS1 & 6.72 & 14.32 & 6.32 & 15.12 & 6.32 & 15.72 & \textbf{11.72} & \textbf{23.88} \\
     DS2 & 7.68 & 15.38 & 6.78 & 14.42 & 6.78 & 18.33 & \textbf{12.68} & \textbf{22.55} \\
     DS3 & 8.72 & 18.23 & 7.72 & 16.36 & 7.72 & 15.32 & \textbf{10.72} & \textbf{18.68} \\
     DS4 & 9.72 & 19.42 & 8.46 & 17.24 & 8.46 & 19.06 & \textbf{13.42} & \textbf{24.33} \\
     DS5 & 7.72 & 14.72 & 6.22 & 15.55 & 6.22 & 15.75 & \textbf{10.72} & \textbf{17.57} \\
     \hline
    \end{tabular}

\end{table}

\begin{figure}[!htbp]
    \centering
 		\includegraphics[width=14cm]{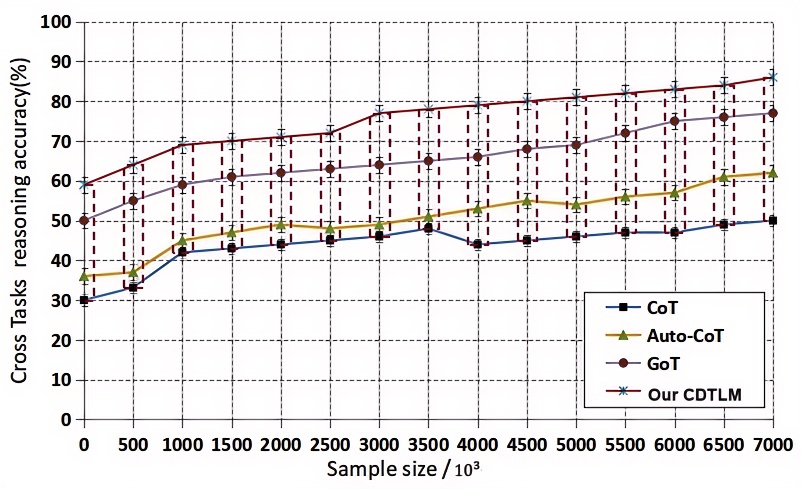}
 	  \caption{Cross-task inference interpretability curves of four algorithms on the DS3 dataset}\label{fig1}
 \end{figure}

\begin{figure}[!htbp]
    \centering
 		\includegraphics[width=14cm]{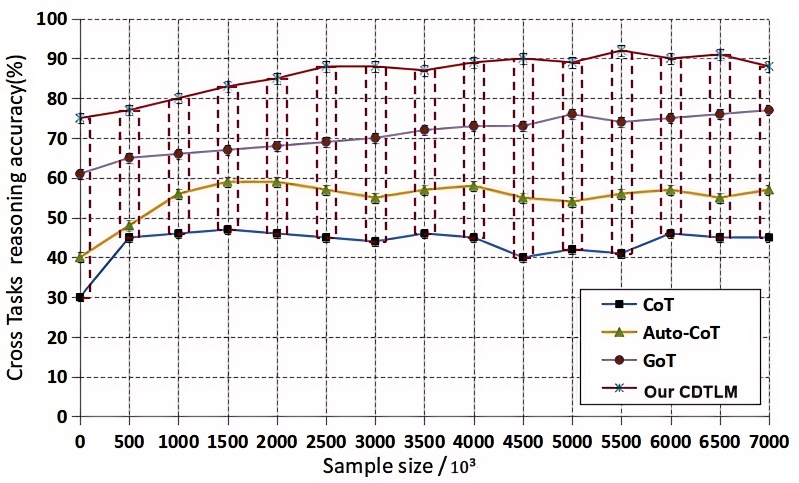}
 	  \caption{Cross-task inference interpretability curves of four algorithms on the DS5 dataset}\label{fig2}
 \end{figure}

Figure \ref{fig1} and \ref{fig2} show the multi-receiver cross-task inference interpretation operation characteristic curve ROC of all algorithms on diverse logically-complex DS3 and DS5 data streams, which can intuitively reflect the explicable advantages. As can be seen from Figure X and Figure X, our proposed CDTLM constructs the mechanism of multi-task interactive interpretation - autonomous cognitive interaction module, which makes this model far better than GoT, Auto-CoT and CoT thought models in the diverse and logically complex DS3 and DS5 data streams. The area under the ROC curve is 10\% and 13\% higher on these two data streams, respectively.

Based on the above experimental results, it can be seen that our proposed CDTLM in this paper, is better than the current mainstream models GoT, Auto-CoT and CoT in terms of accuracy, recall rate, computational efficiency, adaptability and interpretability, and other evaluation indicators, which is due to the improvement of our proposed CDTLM in three aspects. First of all, the definition of CCA can accurately index the industrial subjects and elements in the industrial chain and their dependence relationships, and can realize accurate differentiation and query of real labels with a few types of samples; Secondly, a multi-layer polarization network coding method is proposed to perform cross-layer and cross-domain dynamic association coding integration of large-scale CCAs in each layer. By using the activation principle of hybrid expert model MoE, the thought function domain is constructed, which can support the brain to carry out independent learning, inference, decision-making and regulation.

\subsection{Autonomous decision-making in the resilience loss of the car parts industry chain}
In this part of the experiments, the autonomous decision-making(ADM) task is conducted on the real network data flows, which include five categories: upstream and downstream enterprise value chain DS1, raw material procurement and supply chain DS2, R\&D and production chain DS3, logistics and distribution chain DS4, and sales and after-sales service chain DS5. These experiments aim to evaluate how our proposed ADM can automatically plan resilience adjustment schemes when the industrial chain is attacked by major internal and external events, which is more common in real applications.

\subsubsection{Baselines}
The baseline setting of parameters and system performance experiment of our proposed ADM is compared with the current mainstream algorithm, such as (1) SOAR; (2) ACT-R; (3) OlaGPT et al. To ensure a fair comparison, the four major sets of algorithms chose between five classes of real network data streams and four classes of random fault synthesis data streams, all of which used the same intent-enhancing or voting mechanism. We compare and analyze their task planning completion rate, and task planning success rate. The current mainstream algorithm framework is as follows:

\begin{itemize}
	\item[1)] SOAR\cite{Butt2013soar}: It is designed to support complex problem solving, decision-making, and reasoning. It includes an ontology that describes the state of the world and systems, and a set of rules and operations for reasoning and acting in different situations. It has been used to build a variety of intelligent systems, including expert systems, robot control systems, and learning systems. Its design makes it suitable for tasks that require reasoning in dynamic and uncertain environments.
	\item[2)] ACT-R\cite{Ritter2019act}: It is a rule-based system that combines cognitive models and computer programming techniques to create intelligent systems capable of performing complex tasks. It has been widely used in various fields, including education, human-computer interaction, robot control and artificial intelligence research. It is a powerful tool for studying cognitive processes and has also been used to develop assistive technologies such as educational software and intelligent agents.
	\item[3)] OlaGPT\cite{Xie2023olagpt}: It is a model that simulates the human cognitive processing framework, including attention, memory, inference, learning, and the corresponding scheduling and decision-making mechanisms. Inspired by human active learning, the framework also includes a learning unit to record previous errors and expert opinions, and dynamic reference to improve the ability to solve similar problems.
\end{itemize}

\subsubsection{Evaluation metrics}
This experiment selected real network data flows (including five categories: upstream and downstream enterprise value chain DS1, raw material procurement and supply chain DS2, R\&D and production chain DS3, logistics and distribution chain DS4, and sales and after-sales service chain DS5). From two aspects: task planning completion rate and task execution success rate, our proposed ADM  are compared with SOAR, ACT-R and OlaGPT, and the evaluation metrics are specifically defined as follows:

\begin{itemize}
    \item[1)] Task planning completion rate (TPCR): For a given complex task, the percentage of samples in which the model plans the task and obtains the final answer. This indicator reflects the model's ability to generate executable actions and summarize and refine answers, and is defined as:

    \begin{align}
        TCR=\frac{ {\textstyle \sum_{k=0}^{N}\prod (y_k\ne \phi )} }{N}
    \end{align}
    where ${\textstyle \sum_{k=0}^{N}\prod (y_k\ne \phi )}$ is a judgment function, which means it is 1 when $y_k$ is not empty, otherwise it is 0.

    \item[2)] Task planning success rate (TPSR): For a given question, the proportion of the model outputting the final result and the correct answer. Here, the original data $k$, model output $y_k$, and real label $\hat{y_k}$ are input into the ADM, SOAR, ACT-R, and OlaGPT models, and the model determines whether $_k$ and $\hat{y_k}$ are consistent.

    \begin{align}
        TCR=\frac{ {\textstyle \sum_{k=0}^{N}\prod (y_k = \hat{y_k} )} }{N}
    \end{align}
    where ${\textstyle \sum_{k=0}^{N}\prod (y_k = \hat{y_k} )}$ is a judgment function, which means it is 1 when $y_k$ and $\hat{y_k}$ are equal, otherwise it is 0.
\end{itemize}

\subsubsection{Experimental results and discussions}
Experimental results are shown in Table \ref{perfor1}. Our proposed ADM has the best generalized planning ability and model stability in the whole process of detection accuracy, location accuracy and decision analysis for specific emergencies, no matter on 1.8B or larger data sets, compared with SOAR, ACT-R and OlaGPT. Our proposed ADM Compares with SOAR, ADM and OlaGPT, which the completion rates of task planning in event detection, location and decision making are about 5\%, 5.2\% and 6.1\%, respectively, on all real network data streams. In particular, the decision performance of our proposed ADM  remains relatively stable in both small and large samples.

Table \ref{perfor2} shows the effect of different tasks on all real network data flows. It is further shown that our proposed ADM, which adopts the autonomous loop execution strategy combined with OODA and cognitive CCA, has greatly improved the performance compared with the three models based on SOAR, ACT-R and OlaGPT in terms of task execution success rate in event detection, location and decision, no matter on 1.8B or larger data sets. On all of these real network data streams, our proposed ADM has execution success rates higher than 6\%, 7.3\%, and 6.5\% for event detection, location, and decision making, respectively, compared to SOAR, ACT-R, and OlaGPT models. In particular, the performance of our proposed ADM generation decision strategy remains relatively stable in both small and large samples.

\begin{table}[!htbp]
    \centering

    \caption{Performance comparison of four algorithms planning different task (\%)}
    \label{perfor1}
    \setlength{\tabcolsep}{1mm}
    \renewcommand{\arraystretch}{1.5}
    \scalebox{0.8}{
    \begin{tabular}{c|c|cc|cc|ccc|ccc}
    \hline
     \multirow{2}*{Data Flow}  & \multirow{2}*{Scale(B)} &  \multicolumn{2}{c}{SOAR} & \multicolumn{2}{c}{ACR-R} & \multicolumn{3}{c}{OlaGPT} & \multicolumn{3}{c}{\textbf{Our proposed ADM}}\\
    \cmidrule(lr){3-4}\cmidrule(lr){5-6}\cmidrule(lr){7-9}\cmidrule(lr){10-12}
    & & Observe & Orient & Observe & Orient & Observe & Orient & Decide & Observe & Orient & Decide\\
     \hline
     DS1 & 1.8 & 79.41$\pm$0.31 & 77.91$\pm$0.26 & 75.23$\pm$0.28 & 79.28$\pm$0.32 & 90.24$\pm$0.21 & 91.15$\pm$0.13 &              87.45$\pm$0.31 & 94.11$\pm$0.21 & 95.56$\pm$0.43 & 94.27$\pm$0.31\\
     DS2 & 1.8 & 78.33$\pm$0.21 & 77.88$\pm$0.23 & 75.15$\pm$0.33 & 79.22$\pm$0.14 & 90.12$\pm$0.34 & 92.24$\pm$0.22 &              87.27$\pm$0.21 & 94.21$\pm$0.42 & 95.33$\pm$0.21 & 94.17$\pm$0.24\\
     DS3 & 1.8 & 71.26$\pm$0.12 & 79.32$\pm$0.43 & 73.11$\pm$0.21 & 77.16$\pm$0.51 & 90.17$\pm$0.19 & 88.32$\pm$0.21 &              84.01$\pm$0.08 & 91.13$\pm$0.11 & 92.21$\pm$0.16 & 91.37$\pm$0.21\\
     DS4 & 1.8 & 76.43$\pm$0.34 & 77.26$\pm$0.25 & 75.21$\pm$0.42 & 79.79$\pm$0.21 & 90.33$\pm$0.13 & 89.17$\pm$0.44 &              87.02$\pm$0.52 & 93.43$\pm$0.21 & 95.01$\pm$0.41 & 94.16$\pm$0.63\\
     DS5 & 1.8 & 72.16$\pm$0.21 & 76.23$\pm$0.32 & 72.28$\pm$0.21 & 78.55$\pm$0.41 & 90.44$\pm$0.52 & 87.01$\pm$0.37 &              87.01$\pm$0.11 & 92.01$\pm$0.50 & 93.33$\pm$0.21 & 92.21$\pm$0.78\\
     \hline
    \end{tabular}}

\end{table}

\begin{table}[!htbp]
    \centering

    \caption{Performance comparison of four algorithms planning different task (\%)}
    \label{perfor2}
    \setlength{\tabcolsep}{1mm}
    \renewcommand{\arraystretch}{1.5}
    \scalebox{0.8}{
    \begin{tabular}{c|c|cc|cc|ccc|ccc}
    \hline
     \multirow{2}*{Data Flow}  & \multirow{2}*{Scale(B)} &  \multicolumn{2}{c}{SOAR} & \multicolumn{2}{c}{ACR-R} & \multicolumn{3}{c}{OlaGPT} & \multicolumn{3}{c}{\textbf{Our proposed ADM}}\\
    \cmidrule(lr){3-4}\cmidrule(lr){5-6}\cmidrule(lr){7-9}\cmidrule(lr){10-12}
    & & Observe & Orient & Observe & Orient & Observe & Orient & Decide & Observe & Orient & Decide\\
     \hline
     DS1 & 1.8 & 86.18$\pm$0.33 & 87.98$\pm$0.16 & 85.11$\pm$0.24 & 89.33$\pm$0.43 & 91.24$\pm$0.32 & 86.15$\pm$0.13 &              86.15$\pm$0.31 & 93.17$\pm$0.17 & 92.16$\pm$0.23 & 90.21$\pm$0.34\\
     DS2 & 1.8 & 86.42$\pm$0.24 & 87.56$\pm$0.13 & 85.21$\pm$0.37 & 89.19$\pm$0.32 & 92.12$\pm$0.23 & 86.79$\pm$0.33 &              84.17$\pm$0.41 & 92.21$\pm$0.28 & 91.13$\pm$0.42 & 90.07$\pm$0.41\\
     DS3 & 1.8 & 83.19$\pm$0.17 & 89.21$\pm$0.23 & 83.55$\pm$0.41 & 87.32$\pm$0.19 & 89.67$\pm$0.44 & 84.67$\pm$0.44 &              84.11$\pm$0.18 & 91.33$\pm$0.33 & 90.21$\pm$0.96 & 89.33$\pm$0.21\\
     DS4 & 1.8 & 86.43$\pm$0.34 & 87.16$\pm$0.22 & 85.27$\pm$0.42 & 89.79$\pm$0.21 & 90.43$\pm$0.17 & 86.33$\pm$0.46 &              83.02$\pm$0.43 & 92.43$\pm$0.43 & 93.01$\pm$0.17 & 93.44$\pm$0.43\\
     DS5 & 1.8 & 82.21$\pm$0.17 & 86.23$\pm$0.33 & 82.19$\pm$0.23 & 88.55$\pm$0.24 & 88.22$\pm$0.42 & 83.91$\pm$0.37 &              82.01$\pm$0.19 & 89.01$\pm$0.66 & 88.63$\pm$0.21 & 91.21$\pm$0.66\\
     \hline
    \end{tabular}}

\end{table}

\begin{figure}[!htbp]
    \centering
 		\includegraphics[width=14cm]{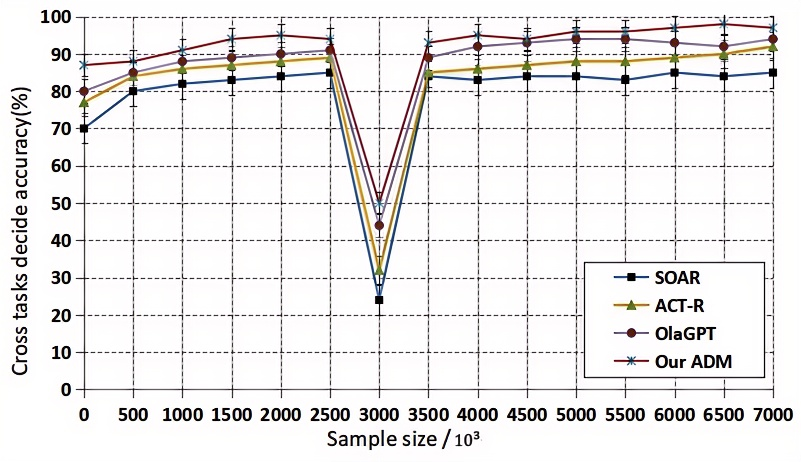}
 	  \caption{Cross-task decision curves of four algorithms on the RD1 mutant data set}\label{fig3}
 \end{figure}

\begin{figure}[!htbp]
    \centering
 		\includegraphics[width=14cm]{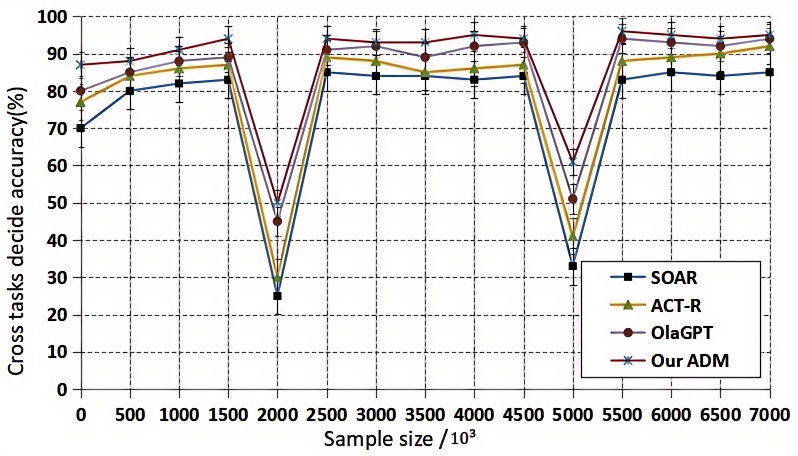}
 	  \caption{Cross-task decision curves of four algorithms on the RD2 repetitive data set}\label{fig4}
 \end{figure}

\begin{figure}[!htbp]
    \centering
 		\includegraphics[width=14cm]{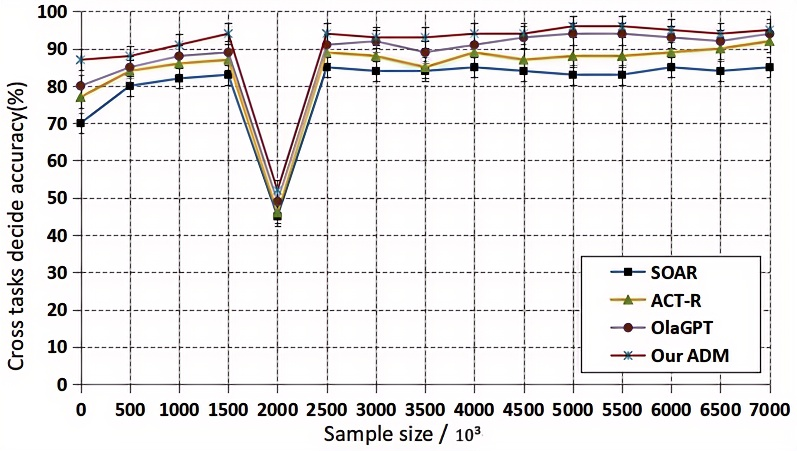}
 	  \caption{Cross-task decision curves of four algorithms on the RD3 incremental data set}\label{fig5}
 \end{figure}

\begin{figure}[!htbp]
    \centering
 		\includegraphics[width=14cm]{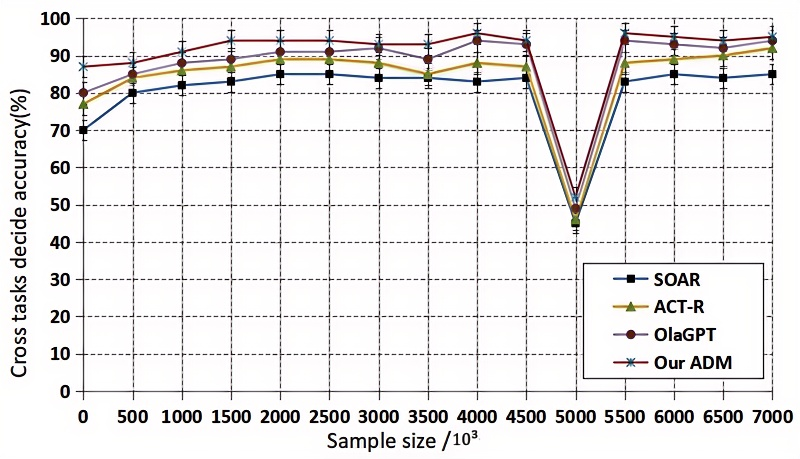}
 	  \caption{Cross-task decision curves of four algorithms on the RD4 gradual data set}\label{fig6}
 \end{figure}

In the execution of tasks such as event detection, location and decision in the environment of four types of random fault synthesis data flow: need balance drift RD1, order and capacity drift RD2, planning and execution drift RD3, logistics and inventory drift RD4. Figure \ref{fig3}, Figure \ref{fig4}, Figure \ref{fig5} and Figure \ref{fig6} show the TPCR of our proposed algorithm and the three comparison algorithms with the increase of sample size on four different types of concept drift data streams. It shows our proposed ADM algorithm has better cross-task TPSR than the three comparison algorithms on 4 different types of event drift data streams. Since the drift widths of mutant and repetitive types are much smaller than those of incremental and gradual types, it can be seen that compared with incremental and gradual types, the decision accuracy rate drops faster when mutation and repetitive types of drift occur, and timely adjustment of training strategies by cognitive thought models can adapt to data drift more quickly. 

Based on the above experimental results, our proposed ADM algorithm is current best cognitive decision-making and planning method. Its superiority is reflected in: 1) the CT-OODA autonomic neural symbol computing framework is defined to realize closed-loop autonomous execution of key links such as event detection, location, decision and regulation in an open environment without manual intervention. 2) the large cognitive thought model provides support and automatic update of various types of rich expert knowledge for the independent cognitive decision-making process, which can improve the accuracy of independent decision-making and regulation. So that it has attention, memory, learning, reasoning, action selection and other independent cognitive decision-making abilities. It can find the operation problems and potential risks of the industrial chain in time from the massive industrial big data, and generate corresponding decision support schemes to help managers complete the optimal allocation and restructuring of industrial chain resources.

\section{Conclusion}\label{5}
The aim of this paper was to present a novel neuro-symbolic autonomous cognitive decision-making and planning framework for handling percolation dynamics problems in giant chaotic network. The proposed framework has advantage on employing neuro-symbolic cognitive diagram of thought, autonomous decision-making, and autonomous planning action into autonomous CT-OODA running engine, where has provides a capability of autonomous cognition, decision, planning for resilience non-equilibrium measurement of industrial chain, facilitating accurate prevention of resilience disasters with linear classifiers. Experiments on 5 datasets and comparisons
with state-of-the-art methods demonstrate the effectiveness
of our proposed methods, which can usually detect vulnerability of the industrial chain network, and autonomously adapt to the diverse evolution of open dynamic network environments. The proposed
framework as "human-in-the-loop" neuro-symbolic hybrid augmented intelligent system in artificial intelligence field. It can be used in various giant high-order interactions network, such as network dynamics modeling, event monitoring, situation assessment, dynamics prediction, catastrophe control. A tool of the proposed framework, called industrial brain, is released for public use.

\section{Acknowledgements}\label{6}
The authors would like to express their sincere gratitude to the anonymous reviewers for their insightful comments and constructive feedback, which greatly contributed to improving the overall quality of this paper. This work was supported in part by the National Key
Research and Development Program of China under Grant 2022YFF0903300; and in part by National Natural Science Foundation of China under Grant 92167109
% Use the style of numbering in square brackets.
% If nothing is used, default style will be taken.
%\begin{enumerate}[a)]
%\item
%\item
%\item
%\end{enumerate}

%\begin{itemize}
%\item
%\item
%\item
%\end{itemize}

%\begin{description}
%\item[]
%\item[]
%\item[]
%\end{description}

% Figure
% \begin{figure}[<options>]
% 	\centering
% 		\includegraphics[<options>]{}
% 	  \caption{}\label{fig1}
% \end{figure}

% \begin{table}[<options>]
% \begin{tabular*}{\tblwidth}{@{}LL@{}}
% \toprule
% \midrule
%  & \\
%  & \\
%  & \\
%  & \\
% \bottomrule
% \end{tabular*}
% \end{table}

% Uncomment and use as the case may be
%\begin{theorem}
%\end{theorem}

% Uncomment and use as the case may be
%\begin{lemma}
%\end{lemma}



\bibliographystyle{cas-model2-names}

\bibliography{cas-refs}



\end{document}